\documentclass[11pt]{article}
\usepackage[preprint]{acl}
\usepackage{times,latexsym}
\usepackage[T1]{fontenc}
\usepackage[utf8]{inputenc}
\usepackage{microtype,inconsolata}
\usepackage{graphicx,amsmath,booktabs,tabularx,array}
\usepackage{placeins}
\newcolumntype{Y}{>{\raggedright\arraybackslash}X}
\graphicspath{{figures/}}
\hypersetup{pdfauthor={Jatin Agrawal and Radhika Mamidi},pdftitle={Presentation Sensitivity in Typed Decision Models for Subjective Text Evaluation},pdfsubject={Research preprint},pdfkeywords={subjective evaluation, typed decision models, presentation sensitivity}}
\title{Presentation Sensitivity in Typed Decision Models\\for Subjective Text Evaluation}
\author{Jatin Agrawal \\
  LTRC, IIITH \\
  \texttt{jatin.agrawal@research.iiit.ac.in}
  \And
  Radhika Mamidi \\
  LTRC, IIITH \\
  \texttt{radhika.mamidi@iiit.ac.in}}

\begin{document}
\maketitle
\begin{abstract}
Typed decision models offer structured choices, scores, and probabilities for low-cost automated evaluation. We present a controlled audit of presentation sensitivity in Jev, Clef Flash, and Perplexity Decider across native choice and score interfaces. Across six datasets with 9,000 multiply annotated humor, helpfulness, and summarization items, we collect 702,000 judgments under interventions that vary positions, identifiers, authorship claims, endorsements, decision framing, and rating rubrics while preserving the evaluated texts. Jev's second-position probability advantage reaches 6.11 percentage points on summaries; reversing reader endorsements changes Perplexity's selected humor text on 40.40\% of pairs; and rejection framing reduces Clef's summary agreement with human preferences by 23.36 points. Rating formats also shift score levels while often preserving rankings. Across these findings, repeatability, presentation robustness, and human alignment emerge as distinct properties of an evaluator. Reliable use therefore requires validating the decisions and score interpretations on which an evaluation pipeline depends.
\end{abstract}

\section{Introduction}
Typed decision models open a route to fast, inexpensive subjective judgment at scale. Jev, released on 15 September 2026 \citep{hardiman2026typesafe}, maps context and typed questions directly to choices, scores, and probabilities. TypeSafe describes a decision-specific architecture trained with Reinforcement Learning for Calibrated Decisions (RLCD), producing probabilities in parallel rather than through autoregressive text generation. Its workflow benchmarks report 444.6$\times$ lower cost and 193.6$\times$ faster execution than the compared LLM configurations \citep{almeida2026jev}. Evaluation tools already document Jev-based response grading \citep{honeyhive2026jev}. As Jev and related models see increasing use in evaluation pipelines, establishing their susceptibility to presentation changes becomes crucial. Jev's decision-only interface provides no generated explanation or chain of thought, making controlled behavioral audits essential for understanding its judgments.

Subjective evaluation requires judging quality amid reasonable human disagreement \citep{plank-2022-problem}. We evaluate three complementary tasks: humor, response helpfulness, and summary quality. For an automated judge, a central reliability question is whether changes in request presentation alter which text is selected, its assigned rating, or agreement with human judgments.

\begin{table*}[t]
\small
\setlength{\tabcolsep}{4pt}
\begin{tabularx}{\textwidth}{@{}>{\raggedright\arraybackslash}p{2.1cm}>{\raggedright\arraybackslash}p{3.05cm}Y Y@{}}
\toprule
Task / interface & Dataset and source & What the model evaluates & Human reference \\
\midrule
Humor / choice & UKP Humor \citep{simpson-etal-2019-predicting} & Funniness of two short texts: puns, jokes, or proverbs/aphorisms & Majority of 5 A/B/neither votes; split cases retained \\
Humor / score & Humicroedit \citep{hossain-etal-2020-semeval} & Funniness of a single-word headline edit, given the original headline & Mean of 5 ratings, 0--3 \\
Helpfulness / choice & HelpSteer2-Preference \citep{ICLR2025_8e237ec6} & Helpfulness of two responses to the same request & Sign of published aggregate preference strength; 5 judgments \\
Helpfulness / score & HelpSteer2 \citep{NEURIPS2024_02fd91a3} & Helpfulness of one response to a request & Mean of 5 helpfulness ratings, 0--4 \\
Summarization / choice & OpenAI comparisons \citep{NEURIPS2020_1f89885d} & Quality of two summaries of the same Reddit post & Majority of 7--23 distinct workers; split cases retained \\
Summarization / score & SummEval \citep{fabbri-etal-2021-summeval} & Relevance of one summary to its source article & Mean of all 8 relevance ratings: 3 expert, 5 crowd; 1--5 \\
\bottomrule
\end{tabularx}
\caption{Three tasks, each evaluated separately through choice and score interfaces. Each dataset contributes 1,500 items, balanced between clearer and more disputed human judgments. Annotation depths describe the selected subsets.}
\label{tab:data}
\end{table*}

LLM-as-a-judge research motivates our tests of position bias, contextual influence, and rating inconsistency \citep{gu2024survey,wang-etal-2024-large-language-models-fair,chen-etal-2024-humans,stureborg2024}. In typed decision models, the key question is how presentation changes propagate through directly returned probabilities, choices, and scores. Measuring these outcomes together reveals whether a shift changes the selected text, affects human alignment, or moves rating levels while preserving rankings.

We evaluate Jev, Clef Flash, and Perplexity Decider on three tasks, each with a pairwise dataset for \texttt{choice} and a rating dataset for \texttt{score}. Interventions preserve evaluated texts; repeated baseline requests establish ordinary variability. Human annotations supply alignment and disagreement references. Three research questions organize the study:

\paragraph{RQ1: Presentation and contextual cues.}
How do candidate positions, identifiers, attributed authorship, endorsements, and evaluator directives influence judgments of unchanged texts?

\paragraph{RQ2: Framing and rating-format sensitivity.}
How do selection versus rejection, an explicit no-preference option, and alternative score rubrics change preferences and ratings?

\paragraph{RQ3: Robustness and human alignment.}
How do these effects relate to repeatability, human disagreement, and alignment, and how consistently do they recur across tasks?

We make four contributions. First, we provide a controlled multi-model audit of native choice and score interfaces across three subjective tasks. Second, factorial and matched-control designs isolate order, identifiers, and cue direction, while paired analyses connect output shifts to decisions and human alignment. Third, we show that repeatability and rank preservation can coexist with consequential presentation sensitivity, yielding concrete validation checks. Fourth, our code and results repository\footnote{\href{https://anonymous.4open.science/r/Presentation-Sensitivity-in-Typed-Decision-Models-for-Subjective-Text-Evaluation-6780/}{Code and results repository (4open.science).}} provides a reproducible framework for evaluating future decision models: preparation, request construction, collection, and analysis scripts; complete aggregate JSON/CSV results; request specifications; and validation tests.

\section{Related Work}
\paragraph{Conventional LLM judges.}
MT-Bench reports strong GPT-4/human agreement alongside position, verbosity, and self-enhancement biases \citep{zheng2023judging}. G-Eval improves human correspondence in summarization and dialogue scoring through chain-of-thought and form filling \citep{liu-etal-2023-g}. Response order can alter GPT-4 and ChatGPT judgments \citep{wang-etal-2024-large-language-models-fair}; JudgeLM addresses position, knowledge, and format biases through training and reference-based techniques \citep{zhu2025judgelm}.

\paragraph{Biases, context, and controls.}
Across 15 LLM judges, position bias varies by judge, task, and candidate-quality gap \citep{shi-etal-2025-judging}. Contextual perturbations, including authority cues, influence both human and LLM judges \citep{chen-etal-2024-humans}. CalibraEval calibrates prediction distributions to reduce position/identifier selection bias \citep{li-etal-2025-calibraeval}, while CALM tests multiple bias families \citep{ICLR2025_fdca08d3}. These findings motivate separate measurements of repetition, order, identifiers, and evaluative context.

\paragraph{Scores and human references.}
JUDGE-BENCH finds that LLM/human alignment varies with task, annotator expertise, and text provenance \citep{bavaresco-etal-2025-llms}. Forced-choice treatment of ambiguous ratings can distort judge validation and selection \citep{guerdan2025indeterminacy}. Together with rating inconsistency \citep{stureborg2024} and human label variation \citep{plank-2022-problem}, these results motivate explicit human references and separate reporting of rating levels, ranks, and disagreement.

\paragraph{Typed decision models.}
The studies above center conventional LLM judges. We investigate whether analogous reliability concerns arise in Jev, Clef Flash, and Perplexity Decider through their native choice and score interfaces. Our audit distinguishes probability and rating shifts, changed decisions, and human alignment, characterizing how presentation sensitivity propagates into subjective evaluation outcomes.

\begin{table*}[tp]
\small
\setlength{\tabcolsep}{4pt}
\begin{tabularx}{\textwidth}{@{}>{\raggedright\arraybackslash}p{2.8cm}rY Y@{}}
\toprule
Choice investigation & Runs & Conditions and controls & Main comparison \\
\midrule
Baseline & 1 & Original question/context; two texts labeled A/B. & Human alignment and reference judgments. \\
Position and A/B identifier bias & 3 & Swap labels, positions, or both; baseline completes the $2\times2$ factorial. & Position, identifier, and interaction effects. \\
Numeric identifier bias & 2 & Fixed text positions labeled 1/2 or 2/1. & Reversed identifier assignments. \\
Authorship attribution & 3 & AI/Human attribution, reversed assignment, unspecified-source control. & Reversal; each assignment versus control. \\
Reader endorsement & 3 & 80\%/20\% reader preference, reversed assignment, unspecified-feedback control. & Reversal; each assignment versus control. \\
Previous-judge feedback & 3 & Selected/not-selected cue, reversed assignment, unspecified-feedback control. & Reversal; each assignment versus control. \\
Explicit no preference & 3 & X first, middle, or last; A/B relative order fixed. & X use/mass, conditional A/B preference, alignment. \\
Selection/rejection framing & 1 & Ask for the worse candidate; complement rejection. & Recovered preference versus baseline. \\
Evaluator-directive injection & 5 & Clean wrapper; inert note or selection directive targeting either candidate. & Directive versus same-target inert note; inert versus clean; clean versus baseline. \\
Repeatability & 4 & Four unchanged repeats beyond baseline. & Decision/probability variation and drift across five calls. \\
\midrule
Total per dataset/model & 28 & \multicolumn{2}{l}{Every condition evaluates all 1,500 selected pairs.} \\
\bottomrule
\end{tabularx}
\caption{Choice experiments, including controls and repeats: 126,000 judgments per model across three datasets. Contrasts measure probability shifts, selected-text changes, and human alignment. Exact questions and cues: Appendices~\ref{app:prompts}--\ref{app:cues}.}
\label{tab:choice_conditions}
\par\medskip
\begin{tabularx}{\textwidth}{@{}>{\raggedright\arraybackslash}p{2.8cm}rY Y@{}}
\toprule
Score investigation & Runs & Conditions and controls & Main comparison \\
\midrule
Baseline & 1 & Original question/context; native rubric (four humor or five helpfulness/relevance levels). & Human alignment and reference ratings. \\
A/B identifier bias & 2 & Candidate-identifier metadata A or B. & A-versus-B ratings of the same text. \\
Numeric identifier bias & 2 & Candidate-identifier metadata 1 or 2. & 1-versus-2 ratings of the same text. \\
Authorship attribution & 3 & AI, Human, or unspecified-source metadata. & AI versus Human; each versus control. \\
Reader endorsement & 3 & High, low, or unspecified reader-rating feedback. & High versus low; each versus control. \\
Previous-judge feedback & 3 & Top-of-scale, bottom-of-scale, or unspecified automated-judge feedback. & High versus low; each versus control. \\
Score-anchor reversal & 1 & Reverse the native verbal anchors. & Polarity-remapped ratings versus baseline. \\
Rating-scale variants & 2 & Three- or seven-level verbal rubrics. & Normalized ratings/ranks versus native rubric; descriptors also vary. \\
Evaluator-directive injection & 3 & Clean wrapper, inert note, or highest-rating directive. & Directive versus inert; inert versus clean; clean versus baseline. \\
Repeatability & 4 & Four unchanged repeats beyond baseline. & Rating variation and drift across five calls. \\
\midrule
Total per dataset/model & 24 & \multicolumn{2}{l}{Every condition evaluates all 1,500 selected texts.} \\
\bottomrule
\end{tabularx}
\caption{Score experiments, including controls and repeats: 108,000 judgments per model across three datasets. Contrasts measure normalized score shifts and alignment changes. Tables~\ref{tab:choice_conditions}--\ref{tab:score_conditions} cover all 702,000 judgments across three models. Exact rubrics and cues: Appendices~\ref{app:prompts}--\ref{app:cues}.}
\phantomsection
\label{tab:score_conditions}
\end{table*}

\section{Study Design and Human References}
\label{sec:data}
\subsection{Tasks, datasets, and references}
Table~\ref{tab:data} describes the six datasets, evaluated qualities, and human references. Pairwise tasks use released comparisons; scoring tasks use the source rating dimension and scale. All datasets come from first-party releases. Appendix~\ref{app:sources} documents source versions, annotation processing, and selection details.

Two reference choices merit emphasis. HelpSteer2 pairwise labels follow the sign of published aggregate preference strength. These overall response preferences serve as a proxy for the helpfulness question asked of the models. SummEval relevance averages all eight individual ratings, weighting expert and crowd judgments by their counts. We additionally examine alignment with each annotator group to assess reference sensitivity (Appendix~\ref{app:results_reference}).

\subsection{Sampling and balancing}
Using a fixed seed, we select 1,500 items per dataset: 750 with clearer human agreement and 750 with more disagreement. Pairwise strata use vote agreement; scoring strata use within-item rating dispersion. UKP balances its humor-category combinations, while scoring subsets retain eligible score-band proportions within agreement strata. Human-preferred candidates are balanced across positions to mitigate position biases in other investigations. Ambiguous pairs are randomly oriented. Selection also limits reuse of source material and the length of HelpSteer2 inputs. Exact thresholds and allocations appear in Appendix~\ref{app:selection}.

\section{Requests and Interventions}
\label{sec:interventions}
\subsection{Choice and score interfaces}
We evaluate Jev \citep{almeida2026jev}, Cloudflare's Clef Flash \citep{chen2026clef}, and Perplexity Decider v1.1 \citep{perplexity2026decider} through OpenRouter's Decisions API \citep{openrouter2026}, using the same request construction for each model. In \texttt{choice} requests, options contain the candidate texts and shared state supplies the task context. For example, humor asks, ``Which option contains the funnier text?'' Table~\ref{tab:prompts} in Appendix~\ref{app:prompts} lists all six instructions and their context fields.

In \texttt{score} requests, shared state contains the candidate and its context, and an ordered list of verbal anchors defines the rating scale. Returned scores use zero-based indices; analysis maps them to the native dataset scale and a common normalized range. Human votes, scores, worker notes, and generating-system identities are withheld from all requests. Model routes, exact anchors, and response handling are documented in Appendices~\ref{app:prompts}--\ref{app:execution}.

Choice outputs contain the selected identifier, option probabilities, and confidence; score outputs contain a scalar rating, rating-level probabilities, confidence, and the scale legend. We analyze the scalar as returned and the probability vectors as reported preferences over the supplied options or levels. Confidence retains its model-specific interpretation. Section~\ref{sec:analysis} defines the derived outcomes.

\subsection{Presentation interventions and controls}
Tables~\ref{tab:choice_conditions} and~\ref{tab:score_conditions} summarize the interventions and their comparisons. Each changes a presentation feature while preserving the evaluated texts. The A/B factorial separates position bias from identifier bias. Position denotes candidate-entry order in the submitted criteria; its contrasts measure submitted-order sensitivity. Attribution and endorsement cues are synthetic metadata: pairwise cues are separated from candidate text, and scoring cues appear in shared state. Each cue family includes an unspecified-cue control to measure effects of the added format itself.

For example, a humor choice presents ``80\% of previous readers preferred this option'' above one unchanged text and ``20\%'' above the other. Shared state instructs: ``Judge only each [TEXT] block for the requested quality. Previous-reader feedback is separate metadata.'' The reversed condition exchanges the two claims. Scoring feedback is supplied as an additional state field. These experiments measure reliance on attributed information. Appendix~\ref{app:request_examples} gives complete request examples.

Framing interventions change how a judgment is requested. The no-preference condition distinguishes abstention from redistribution between A and B; its placement is reported separately. Selection/rejection framing compares the preferred text recovered from asking for the better versus worse candidate. Rating-format variants reverse the native anchors or introduce three-/seven-level verbal descriptions; remapped scores provide a common comparison scale. Exact wording, directive controls, and mappings appear in Appendix~\ref{app:cues}.

\FloatBarrier
\section{Measurement and Inference}
\label{sec:analysis}
\subsection{Outcomes and comparison units}
Each intervention is compared with its designated control on matched items. Movement analyses and scoring alignment use all 1,500 items per dataset; pairwise human alignment uses the subset with a strict A/B human preference. Table~\ref{tab:results_baseline} gives those subset sizes.

\paragraph{Pairwise outcomes.}
We map choices back to the underlying texts to separate position and identifier effects. Outcomes are the mean signed candidate-probability shift, the proportion of pairs whose selected text changes, and the change in agreement with the human preference. Probability-based analyses renormalize returned vectors to unit mass. For no-preference conditions, we report X selection and probability mass alongside conditional A/B preference.

\paragraph{Factorial contrasts.}
For fixed candidates $u,v$, let $p_0,p_L,p_P,p_{LP}$ denote $P(u)$ under baseline, exchanged labels, exchanged positions, and both exchanges. The main effects are
\begin{align}
d_{\mathrm{first}}&=(p_0+p_L-p_P-p_{LP})/2,\\
d_{\mathrm{B}}&=(p_L+p_{LP}-p_0-p_P)/2.
\end{align}
Positive values favor first position or label B, respectively; each averages over the other factor. An interaction tests whether identifier bias depends on position (Appendix~\ref{app:cues}, Eq.~\ref{eq:factorial_interaction}). The same contrasts apply to selected-text indicators. Attribution and endorsement contrasts compare reversed cue assignments within the same metadata format.

\paragraph{Scoring outcomes.}
Primary scoring analyses use the returned scalar score. For $K$ ordered levels, score $s$ maps to $s/(K-1)$, or $1-s/(K-1)$ under reversed anchors. Human means on the native dataset scale $[L,U]$ map to $(h-L)/(U-L)$. Probability and normalized-score shifts are reported in percentage points. Human alignment uses MAE, RMSE, signed error, and Pearson/Spearman correlations. Scores reconstructed from normalized rating probabilities provide sensitivity checks (Appendix~\ref{app:results_reference}). Score compression uses dispersion and score-versus-human relationships; observational length associations adjust for human-reference and task covariates.

\paragraph{Repeatability.}
The canonical successful baseline and four unchanged repeats measure within-item variation. The item remains the comparison unit. Appendix~\ref{app:execution} describes baseline reconstruction and response validation; Appendix~\ref{app:results_humans} gives repeatability and human-reference diagnostics.

\subsection{Uncertainty and multiple comparisons}
Confidence intervals account for dependence from reused content. We cluster by source context, article, prompt, or headline as appropriate. UKP pairs can share either joke, so we use a shared-joke dyadic variance estimator \citep{aronow2015}. Paired mean differences use Student-$t$ intervals, with cluster or joke-node counts determining degrees of freedom. Seeded cluster bootstrap checks assess sensitivity. Appendix~\ref{app:statistics} gives the estimators and diagnostics.

Effect and association tests are two-sided, with discoveries defined by Benjamini--Yekutieli (BY) adjusted $q<.05$ \citep{benjamini2001}. We report effect sizes and pointwise 95\% confidence intervals. The primary BY family contains 369 tests of signed probability, selection, and scoring outcomes, including baseline alignment differences and between-model intervention contrasts. A separate secondary family contains 1,656 tests of non-repeat condition/control effects, human alignment/error changes, susceptibility correlations, and pooled effects. A comparison can supply tests for several outcomes; Appendix~\ref{app:multiplicity} gives family membership. Subgroup summaries, absolute movement, and sensitivity diagnostics are reported descriptively.

\subsection{Practical consequences}
We assess whether interventions change decisions or human alignment. The 450 condition/control comparisons summarize signed and absolute movement, changed selections, and alignment/error changes, including unchanged-repeat controls. Non-repeat signed outcomes enter the secondary testing family (Appendix~\ref{app:multiplicity}). Supplementary analyses examine shared susceptibility and cross-task heterogeneity; their methods and results appear in Appendices~\ref{app:susceptibility}--\ref{app:pooling} and~\ref{app:results_synthesis}.

\section{Results}
\label{sec:results}
Baseline calls are highly repeatable, while intervention effects vary by presentation feature, interface, model, and task. We report choice and score findings separately and connect probability or rating shifts to changed decisions and human alignment. Clef Flash and Perplexity Decider are abbreviated as Clef and Perplexity.

\begin{table*}[t]
\centering\footnotesize
\setlength{\tabcolsep}{3.5pt}
\begin{tabular}{@{}lrrrrr@{}}
\toprule
\shortstack[l]{Task\\{}model} & \shortstack[l]{Pair alignment\\{}\%} & \shortstack[l]{Pair repeat\\{}flip \%} & \shortstack[l]{Score\\{}MAE} & \shortstack[l]{Score\\{}Spearman} & \shortstack[l]{Score repeat\\{}abs. delta pp} \\
\midrule
Humor / Jev & 73.38 & 3.15 & 0.256 & 0.365 & 1.09 \\
Humor / Clef & 67.45 & 3.02 & 0.200 & 0.549 & 1.42 \\
Humor / Perplexity & 76.76 & 0.08 & 0.158 & 0.379 & 0.00 \\
Helpfulness / Jev & 68.62 & 2.58 & 0.175 & 0.448 & 0.89 \\
Helpfulness / Clef & 61.85 & 3.20 & 0.184 & 0.271 & 1.06 \\
Helpfulness / Perplexity & 73.53 & 0.00 & 0.148 & 0.581 & 0.00 \\
Summary / Jev & 78.98 & 1.70 & 0.132 & 0.384 & 0.86 \\
Summary / Clef & 74.43 & 2.65 & 0.106 & 0.368 & 1.04 \\
Summary / Perplexity & 77.05 & 0.00 & 0.115 & 0.343 & 0.00 \\
\bottomrule
\end{tabular}
\caption{Baseline human alignment and unchanged-request variation. Pair alignment uses strict A/B references only (UKP n=1,063; HelpSteer2 n=1,447; OpenAI n=1,451); score MAE uses normalized 0-1 human means (n=1,500). Repeat columns average four comparisons with the canonical baseline on the same items. Choice and score results use the datasets in Table 1. Values are descriptive and rounded.}
\label{tab:results_baseline}
\end{table*}

\subsection{Choice judgments}

\paragraph{Repeatability and reference agreement.}
Perplexity has the highest baseline agreement on humor and helpfulness, while Jev leads on summaries (Table~\ref{tab:results_baseline}). Unchanged-request choice flips average 1.70--3.20\% for Jev/Clef and at most 0.08\% for Perplexity.

\begin{table*}[t]
\centering\footnotesize
\setlength{\tabcolsep}{3.5pt}
\begin{tabular}{@{}lrrr@{}}
\toprule
\shortstack[l]{Task\\{}contrast} & \shortstack[l]{Jev\\{}pp [95\% CI]} & \shortstack[l]{Clef Flash\\{}pp [95\% CI]} & \shortstack[l]{Perplexity\\{}pp [95\% CI]} \\
\midrule
Humor: First position & -2.3 [-3.0, -1.5]* & 0.0 [-0.3, 0.3] & 4.6 [4.1, 5.1]* \\
Humor: B versus A & 3.9 [3.7, 4.1]* & -8.6 [-9.6, -7.7]* & -1.5 [-1.7, -1.3]* \\
Humor: Human versus AI & 7.0 [6.7, 7.4]* & 3.6 [3.1, 4.1]* & 9.0 [8.6, 9.4]* \\
Humor: Reader feedback & 13.5 [13.0, 14.1]* & 4.8 [4.4, 5.2]* & 27.8 [26.8, 28.8]* \\
Helpfulness: First position & -2.3 [-2.9, -1.6]* & -0.1 [-0.3, 0.0] & -4.4 [-4.8, -3.9]* \\
Helpfulness: B versus A & -1.6 [-1.8, -1.4]* & -4.7 [-5.2, -4.1]* & -0.4 [-0.6, -0.3]* \\
Helpfulness: Human versus AI & 3.4 [3.1, 3.6]* & -0.2 [-0.5, 0.0] & 3.8 [3.6, 4.0]* \\
Helpfulness: Reader feedback & 13.7 [13.2, 14.2]* & 4.4 [4.2, 4.7]* & 19.1 [18.6, 19.5]* \\
Summary: First position & -6.1 [-6.9, -5.3]* & 0.0 [-0.2, 0.2] & -0.4 [-0.9, 0.1] \\
Summary: B versus A & -0.5 [-0.7, -0.4]* & 8.2 [7.2, 9.3]* & -1.4 [-1.6, -1.2]* \\
Summary: Human versus AI & 13.7 [13.0, 14.3]* & 32.1 [31.0, 33.3]* & 7.0 [6.7, 7.3]* \\
Summary: Reader feedback & 12.7 [12.1, 13.4]* & 7.1 [6.7, 7.6]* & 22.6 [22.1, 23.2]* \\
\bottomrule
\end{tabular}
\caption{Choice probability effects on unchanged texts, $n=1{,}500$ per task/model. Positive values favor first position, label B, Human attribution, or favorable reader feedback. Position and B effects average over the other factorial factor; cue effects reverse both assignments within the same format. Intervals are pointwise and dependence-aware; * marks primary-family BY $q<.05$. Complete identifiers and cue controls: Appendices~\ref{app:results_identifiers}--\ref{app:results_cues}.}
\label{tab:results_choice_effects}
\end{table*}

\paragraph{Position and identifier effects.}
Jev favors second position across all tasks; its first-position effect reaches -6.11 [-6.95, -5.26] pp on summaries (Table~\ref{tab:results_choice_effects}). Clef's position effects remain close to zero, while its B-label preference reverses from humor to summaries. Perplexity reverses its position preference between humor and helpfulness. Jev favors numeric identifier 2 across tasks, with a 7.44 pp humor effect; the other models vary in direction. Position--identifier interactions are at most 0.30 pp and none passes correction. Complete results appear in Appendix~\ref{app:results_identifiers}.

These effects reach discrete decisions: on summaries, Jev's first-position selection contrast is -8.73 [-10.42, -7.05] pp, and Clef's B-label selection contrast is 13.30 [11.20, 15.40] pp. Both are signed factorial contrasts, averaging over the other factor (Table~\ref{tab:appendix_factorial_choice}).

\paragraph{Authorship and feedback sensitivity.}
Human attribution generally increases candidate preference, most strongly for Clef summaries: 32.15 [31.02, 33.28] pp, with 39.53\% changed selections. Reader endorsements consistently favor the endorsed candidate: Jev's effects span 12.71--13.71 pp and Perplexity's 19.07--27.80 pp. Previous-judge effects are smaller; Jev's summary effect is 0.28 pp ($q=0.088$), while the other configurations pass correction (Appendix~\ref{app:results_cues}).

The format itself also matters: the unspecified-source control shifts Jev's summary probability for original candidate A by -4.72 [-5.20, -4.24] pp versus baseline. Reversing attributed sources within the same format separates the attribution contrast from this format-control effect.

\begin{table*}[t]
\centering\footnotesize
\setlength{\tabcolsep}{3.5pt}
\begin{tabular}{@{}lrrr@{}}
\toprule
\shortstack[l]{Task\\{}outcome} & \shortstack[l]{Jev} & \shortstack[l]{Clef Flash} & \shortstack[l]{Perplexity} \\
\midrule
Humor: reader reversal (\%) & 22.00 & 10.13 & 40.40 \\
Humor: unchanged repeat (\%) & 3.15 & 3.02 & 0.08 \\
Humor: rejection agreement (pp) & -1.2 [-3.3, 0.8] & -2.1 [-5.5, 1.3] & 0.9 [-0.9, 2.7] \\
Helpfulness: reader reversal (\%) & 18.27 & 9.73 & 27.13 \\
Helpfulness: unchanged repeat (\%) & 2.58 & 3.20 & 0.00 \\
Helpfulness: rejection agreement (pp) & -1.2 [-3.1, 0.7] & -15.8 [-19.9, -11.6]* & -1.1 [-2.7, 0.5] \\
Summary: reader reversal (\%) & 16.20 & 12.47 & 34.20 \\
Summary: unchanged repeat (\%) & 1.70 & 2.65 & 0.00 \\
Summary: rejection agreement (pp) & -4.5 [-6.8, -2.3]* & -23.4 [-27.0, -19.7]* & 1.7 [-0.1, 3.4] \\
\bottomrule
\end{tabular}
\caption{Practical choice consequences. Reader rows give selected-text changes under reversed endorsements; repeat rows average four unchanged-call comparisons with baseline (both descriptive \%, all 1,500 pairs). Rejection rows give recovered-rejection minus baseline human agreement in pp, with pointwise 95\% CIs; * marks secondary-family BY $q<.05$. Strict reference counts: 1,063/1,447/1,451 for humor/helpfulness/summary. Cue-versus-control agreement changes: Table~\ref{tab:appendix_reader_alignment}.}
\label{tab:results_choice_consequences}
\end{table*}

\begin{table*}[t]
\centering\footnotesize
\setlength{\tabcolsep}{3.5pt}
\begin{tabular}{@{}lrrr@{}}
\toprule
\shortstack[l]{Task\\{}contrast} & \shortstack[l]{Jev\\{}pp [95\% CI]} & \shortstack[l]{Clef Flash\\{}pp [95\% CI]} & \shortstack[l]{Perplexity\\{}pp [95\% CI]} \\
\midrule
Humor: Reader feedback & 4.0 [3.8, 4.1]* & 7.3 [7.0, 7.7]* & 21.2 [20.8, 21.6]* \\
Humor: Previous judge & 2.5 [2.4, 2.5]* & 2.1 [1.9, 2.3]* & 19.1 [18.8, 19.4]* \\
Humor: Three levels & -2.8 [-2.9, -2.7]* & -3.5 [-3.8, -3.3]* & -6.8 [-6.9, -6.7]* \\
Humor: Seven levels & -7.6 [-7.8, -7.5]* & -5.2 [-5.4, -5.0]* & -3.7 [-4.0, -3.5]* \\
Helpfulness: Reader feedback & 16.5 [16.1, 16.8]* & 3.4 [3.3, 3.6]* & 10.3 [10.1, 10.4]* \\
Helpfulness: Previous judge & 11.8 [11.5, 12.0]* & 1.6 [1.4, 1.7]* & 9.3 [9.1, 9.4]* \\
Helpfulness: Three levels & 2.2 [2.0, 2.4]* & 4.7 [4.4, 5.0]* & 0.8 [0.6, 0.9]* \\
Helpfulness: Seven levels & 1.4 [1.3, 1.5]* & -1.7 [-1.8, -1.5]* & 0.8 [0.7, 0.9]* \\
Relevance: Reader feedback & 5.7 [5.4, 5.9]* & 3.0 [2.6, 3.3]* & 6.9 [6.8, 7.0]* \\
Relevance: Previous judge & 5.2 [5.0, 5.4]* & 0.8 [0.7, 1.0]* & 10.3 [10.1, 10.4]* \\
Relevance: Three levels & 5.9 [5.6, 6.1]* & 7.9 [7.6, 8.2]* & 1.1 [0.9, 1.4]* \\
Relevance: Seven levels & 2.6 [2.5, 2.7]* & -0.7 [-0.9, -0.6]* & 0.7 [0.6, 0.8]* \\
\bottomrule
\end{tabular}
\caption{Score effects in normalized-score pp, $n=1{,}500$ per task/model. Feedback compares favorable with unfavorable claims; scale variants compare with the native rubric. Positive values indicate higher ratings. Alternative formats change descriptions as well as level count. Pointwise 95\% CIs; * marks primary-family BY $q<.05$. Rank associations, anchor reversal, and reference-error changes: Appendix~\ref{app:results_formats}.}
\label{tab:results_score_effects}
\end{table*}

These shifts have observable decision consequences (Table~\ref{tab:results_choice_consequences}). Reversing reader endorsements changes Perplexity's selected humor text on 40.40\% of pairs. Relative to unspecified feedback, assigning high feedback to the first or second candidate reduces its humor agreement by 4.6 and 6.0 pp, respectively; both pass secondary correction. Other models and tasks have smaller or uncertain net agreement changes despite changed choices (Table~\ref{tab:appendix_reader_alignment}).

\paragraph{Framing, directives, and no preference.}
Rejection framing changes recovered preferences in every configuration. Clef's helpfulness agreement falls from 61.85\% to 46.10\%, and its summary agreement changes by -23.36 [-27.01, -19.72] pp. On helpfulness, Clef selects the same text as both more and less helpful on 67.73\% of all pairs, consistent with incomplete adaptation to question polarity. Perplexity's changes have little supported net effect on agreement. Directive effects are selective: its second-target summary directive raises target probability by 10.14 [9.69, 10.59] pp versus an inert note, whereas Jev's effects are mostly small (Appendix~\ref{app:results_injection}).

With X last, UKP neither-majority cases yield 43/187 X choices for Jev, 19/187 for Clef, and 88/187 for Perplexity. All models choose X on 0/53 HelpSteer aggregate-neutral cases and 1/49 OpenAI split-vote cases. These distinct reference groups and the placement-specific changes are detailed in Appendix~\ref{app:results_framing}. Length and human-disagreement associations are task-specific descriptive results (Appendix~\ref{app:results_humans}).

\subsection{Scoring judgments}

\paragraph{Feedback changes rating levels.}
Reader and previous-judge feedback raise scores in every configuration (Table~\ref{tab:results_score_effects}). Jev's helpfulness reader contrast is 16.47 [16.12, 16.81] normalized-score pp; Perplexity's largest feedback effects occur on humor. Error changes depend on the reference and starting scores. For example, favorable reader feedback increases Perplexity's humor MAE by 4.8 pp relative to unspecified feedback, while unfavorable feedback has little net effect. All cue-versus-control error changes appear in Table~\ref{tab:appendix_reader_error}.

\paragraph{Rating formats affect levels and ranks differently.}
Jev's seven-level humor format lowers ratings by 7.63 pp, while its three-level relevance format raises them by 5.87 pp. Its rank correlations with native-format scores remain 0.963--0.996. Clef relevance is less rank-stable, with correlations of 0.903--0.938 across formats. Thus high rank association can coexist with shifts in absolute ratings, with the extent of rank preservation varying by configuration. These level shifts matter for judge design: fixed thresholds and aggregate quality scores depend on rating levels even when candidate rankings remain similar. Reversed anchors also shift remapped ratings; full estimates, error changes, and the format figure appear in Appendix~\ref{app:results_formats}.

\paragraph{Identifiers, attribution, and directives are smaller.}
Scoring A/B, numeric, and authorship effects reach at most 1.35, 0.94, and 1.57 pp in magnitude. Several small shifts pass correction. Directive effects are mixed: Clef relevance falls by -2.11 [-2.44, -1.79] pp despite a highest-rating instruction; Perplexity relevance rises by 2.74 pp while its MAE decreases by 1.43 pp. These cases separate movement from distance to the specified human mean (Appendices~\ref{app:results_identifiers}--\ref{app:results_injection}).

\paragraph{Score dispersion and reference dependence.}
Judge-score dispersion relative to human ratings varies by dataset: Jev's judge/human SD ratio is 0.58 for humor and 1.10 for relevance. On SummEval, its Pearson correlation is 0.597 with experts, 0.061 with crowd raters, and 0.382 with the pooled reference (Appendices~\ref{app:results_formats}--\ref{app:results_reference}). Score repeat variation is small for Jev/Clef and zero for Perplexity (Table~\ref{tab:results_baseline}).

Cross-task pools, shared-susceptibility associations, and robustness checks appear in Appendices~\ref{app:results_synthesis}--\ref{app:results_robustness}.
\label{sec:results_end}
\FloatBarrier

\section{Discussion}
\paragraph{Order and identifiers are model-specific.}
Jev favors the second position across all three choice tasks; Clef remains near zero, and Perplexity changes direction by task. Identifier preferences vary and position--identifier interactions remain small. LLM-judge studies likewise find that order effects depend on the judge, task, and candidate-quality gap \citep{wang-etal-2024-large-language-models-fair,shi-etal-2025-judging}. Jev alone shows a uniform serial-order tendency across our tasks. On the score interface, A/B, numeric, and authorship effects are at most 1.57 pp, much smaller than evaluative-cue and rubric effects.

\paragraph{Endorsements are the common susceptibility.}
Reader endorsements move choice probabilities toward the endorsed text for every model and task; reader and previous-judge feedback also raise scores in every configuration. At their strongest, endorsement-driven choice reversals far exceed unchanged-request variation. This unanimous cross-model effect extends the authority and context sensitivity reported for conventional LLM judges \citep{chen-etal-2024-humans}. Matched controls attribute the principal shifts to cue direction rather than added metadata.

\paragraph{Framing and rating formats change downstream use.}
Clef frequently chooses the same helpfulness candidate as both better and worse, while rejection framing sharply reduces its summary alignment. Together with evidence that forced-choice handling can distort LLM-judge validation under rating ambiguity \citep{guerdan2025indeterminacy}, this shows that decision framing is part of evaluator validity. Score-format variants shift rating levels despite high within-model rank correlations, echoing rating inconsistency in LLM evaluators \citep{stureborg2024}. Rank-based comparisons may therefore remain stable while thresholds and mean scores change.

\paragraph{Reliability has distinct dimensions.}
Perplexity nearly reproduces unchanged calls yet shows the largest endorsement-driven reversals; Jev's SummEval alignment differs substantially between expert and crowd references. As in LLM-judge research \citep{shi-etal-2025-judging,bavaresco-etal-2025-llms}, repeatability, presentation robustness, and human alignment answer different questions.

\section{Conclusion}
Typed decision models are inexpensive and structured, yet presentation-sensitive. Across three models and three tasks, endorsements are the most consistent influence; order and identifiers vary by model; and framing and rubrics alter decisions or ratings despite stable repeats and ranks. Deployment audits should vary order and identifiers independently, control contextual cues, check framing consistency, and compare score levels with ranks.

\label{sec:main_end}

\section*{Limitations}
The study compares three typed decision models across three English-language tasks. Comparisons with conventional LLM judges use published evidence; future work can evaluate both model families under matched conditions. Sampling balances agreement strata and UKP humor categories to cover a range of evaluation cases. Human references are fixed aggregates, and incomplete worker identities leave some shared-rater dependence unresolved. SummEval combines expert and crowd ratings, while HelpSteer2 overall preferences serve as a proxy for helpfulness. Requests use adapted single-item questions, and structural selection retains examples with potentially incomplete context.

Effects characterize the implemented wording, datasets, and cue placement. Alternative rubrics change descriptions, anchor reversal changes numeric polarity, and inserting X first or middle changes candidate positions. Directive controls have approximately matched lengths. Linear score normalization assumes comparable spacing between ordinal levels. The three selected tasks restrict cross-task inference, and available annotations represent human-reference uncertainty only partially.

\section*{Ethical Considerations}
We reuse released annotations and follow upstream attribution and data-use terms. Content may include offensive humor or sensitive requests. Authorship and feedback cues are fabricated solely for controlled presentation experiments. Reproducibility materials preserve their experimental status and respect source redistribution restrictions.

\bibliography{bibliography}
\clearpage
\appendix
\section{Data Sources and Selection Details}
\label{app:sources}
The following first-party releases supply the six datasets. The accompanying \texttt{source\_evidence.json} records revisions and frozen input hashes.

\paragraph{UKP Humor.}
\href{https://github.com/UKPLab/acl2019-GPPL-humour-metaphor/tree/5f2a432c4db5191467a3315220b78c926f66f6b3}{Official UKP release}, revision \texttt{5f2a432c4db5}.
The sampling frame contains 28,210 pairs across heterographic puns, homographic puns, non-pun jokes, and non-humorous proverbs/aphorisms. Five annotators choose A, B, or neither; three matching votes define a majority. Equal allocation across ten unordered category combinations yields 750 candidate appearances per category. Joke IDs identify reused texts.

\paragraph{Humicroedit.}
\href{https://zenodo.org/records/3969509}{Author-deposited Zenodo release 3969509}.
Subtask 1 train, development, and test files are pooled, requiring exactly five grades. Released single-word replacements reconstruct edited headlines; selection retains at most one edit per original headline.

\paragraph{HelpSteer2.}
\href{https://huggingface.co/datasets/nvidia/HelpSteer2/tree/990b2711a36180dd19d9c94b8627844866f8982a}{Official NVIDIA release}, revision \texttt{990b2711a361}.
Choice items use the preference file; scoring items join exact train/validation prompt--response strings to the raw disagreement release. Both require five judgments. Pairwise references follow published preference-strength sign, with zero classified as aggregate-neutral. Scoring references average five raw helpfulness ratings; published integers support sampling. Strength sign and raw directional majority can differ.

\paragraph{OpenAI summary comparisons.}
The \href{https://github.com/openai/summarize-from-feedback#human-feedback-data}{official dataset documentation} identifies the \href{https://openaipublic.blob.core.windows.net/summarize-from-feedback/dataset/comparisons/}{first-party comparison files}.
Only records containing Reddit posts enter the pairwise sampling frame. Pair order is canonicalized before worker-level consolidation. Consistent repeated worker votes count once; conflicting worker--pair votes are excluded, with at least seven distinct workers remaining. Context IDs use post text; the lexicographically first released title variant, if any, is prepended to the post. The selected 1,500 pairs cover 540 contexts, with at most three pairs per context.

\paragraph{SummEval.}
The \href{https://github.com/Yale-LILY/SummEval}{official repository} links the \href{https://storage.googleapis.com/sfr-summarization-repo-research/model_annotations.aligned.jsonl}{aligned annotation release}.
The \href{https://cs.nyu.edu/~kcho/DMQA/}{original story archives} supply articles via released file paths, retaining content before the first \texttt{@highlight} and joining nonempty lines. Selected records cover 100 articles and contain three expert and five crowd relevance ratings. The reference is $(3\bar h_{\mathrm{expert}}+5\bar h_{\mathrm{crowd}})/8$, assigning equal weight to each individual rating. Exact article--summary duplicates retain one annotation record.

\subsection{Selection strata}
\label{app:selection}
Structural eligibility requires the relevant text fields and annotation depth. Sampling uses a fixed seed, stable item IDs, and selection without replacement. Every dataset contributes 750 clear and 750 disputed items. Clear pairwise items require 4/5 agreement on A/B/neither for UKP, 4/5 raw directional agreement for HelpSteer2, or majority share $\geq .80$ for OpenAI. Remaining pairs are disputed. HelpSteer2's direction-based agreement stratum can differ from its strength-sign reference.

Scoring strata split at the eligible-pool median population SD, with equality assigned to the clear group. SummEval dispersion is the mean of within-expert and within-crowd SDs. This defines relative agreement within each dataset. UKP contributes 150 pairs per unordered category combination, 75 per agreement stratum.

Within each agreement group, Humicroedit mean-score bands are $[0,1)$, $[1,2)$, and $[2,3]$; HelpSteer2 uses published integer levels 0--4; SummEval uses mean-relevance bands $[1,2)$, $[2,3)$, $[3,4)$, and $[4,5]$. Largest-remainder allocation preserves eligible-pool band proportions subject to integer quotas and reuse constraints. HelpSteer2's analysis reference remains the raw five-rating mean. Stable per-source pseudorandom streams derive from the common seed and source-name hashes. Sample IDs, allocation cells, orientation decisions, annotation arrays, and source-file hashes are retained in the prepared records and manifest.

HelpSteer2 limits prompt/history plus candidate responses to 5,000 words, counting shared context once, and selects one scored response per prompt. Inputs remain complete; eligibility checks fields and annotation depth. Strict A/B winners are balanced within category/agreement strata, with odd remainders compensated to keep total position imbalance at most one. Ambiguous pairs are randomly oriented. Baseline labels are A then B.

\section{Exact Baseline Request Specification}
\label{app:prompts}
The request envelope is \texttt{\{model, state, questions\}}, with one \texttt{questions.judgment} entry containing \texttt{type}, \texttt{instructions}, and \texttt{criteria}. Pairwise criteria map labels to full texts; scoring criteria list ordered verbal anchors. Shared state supplies task context, including conversation history where present.

\begin{table*}[t]
\small
\begin{tabularx}{\textwidth}{@{}>{\raggedright\arraybackslash}p{2.1cm}>{\raggedright\arraybackslash}p{3.05cm}Y@{}}
\toprule
Component & Shared state & Exact instructions \\
\midrule
UKP Humor & Two standalone short texts are being compared. & Which option contains the funnier text? \\
Humicroedit & \texttt{original\_headline}, \texttt{edited\_headline} & How funny is \texttt{`edited\_headline`} as an edit of \texttt{`original\_headline`}? \\
HelpSteer2 pairs & \texttt{user\_request} & Which option contains the more helpful response to \texttt{`user\_request`}? \\
HelpSteer2 scores & \texttt{user\_request}, \texttt{response} & How helpful is \texttt{`response`} overall for \texttt{`user\_request`}? \\
OpenAI pairs & \texttt{source\_text} & Which option contains the better summary of \texttt{`source\_text`}? \\
SummEval & \texttt{source\_article}, \texttt{summary} & How relevant is \texttt{`summary`} to \texttt{`source\_article`}? Relevance means selecting important content from the source. Penalize redundant or excess information. \\
\bottomrule
\end{tabularx}
\caption{Exact baseline instructions and context fields. Backticks around field names are literal characters. Scoring requests place the evaluated text in shared state.}
\label{tab:prompts}
\end{table*}

\paragraph{Humor anchors (indices 0--3).}
``Not funny.''; ``Slightly funny.''; ``Moderately funny.''; ``Funny.''

\paragraph{Helpfulness anchors (indices 0--4).}
The ordered descriptions are:
\begin{enumerate}
\item Not useful or helpful at all; completely misses the essence of the user's request.
\item Borderline unhelpful; mostly does not capture what the user wanted, although it may still be usable in a small way.
\item Partially helpful; addresses the request but misses the overall goal in some way, so it does not fully satisfy the user.
\item Mostly helpful; mainly aligned with the user's intent and largely satisfies the request, with some room for improvement.
\item Extremely helpful; completely aligned with the spirit of the user's request.
\end{enumerate}
The displayed list enumerates five descriptions; API indices start at zero.

\paragraph{Relevance anchors (indices 0--4; native 1--5).}
``Very low relevance.''; ``Low relevance.''; ``Moderate relevance.''; ``High relevance.''; ``Very high relevance.'' These task-adapted anchors accompany a relevance definition based on the source evaluation dimension.

\subsection{Complete feedback-request examples}
\label{app:request_examples}
These literal payloads are reconstructed offline from selected records using the collection runner. They include all submitted fields; the first assigns high feedback to original A and low feedback to B. Scoring adds feedback as shared-state metadata. References and internal analysis mappings are withheld.
\paragraph{ukp\_humor\_pairwise.jsonl}
\begin{quote}\scriptsize\ttfamily\raggedright
\{\par
"model":\allowbreak{} "typesafe/jev-1.13",\par
"state":\allowbreak{} \{\par
"task\_context":\allowbreak{} "Two\allowbreak{} standalone\allowbreak{} short\allowbreak{} texts\allowbreak{} are\allowbreak{} being\allowbreak{} compared.",\par
"experimental\_format":\allowbreak{} "Judge\allowbreak{} only\allowbreak{} each\allowbreak{} [TEXT]\allowbreak{} block\allowbreak{} for\allowbreak{} the\allowbreak{} requested\allowbreak{} quality.\allowbreak{} Previous-reader\allowbreak{} feedback\allowbreak{} is\allowbreak{} separate\allowbreak{} metadata."\par
\},\par
"questions":\allowbreak{} \{\par
"judgment":\allowbreak{} \{\par
"type":\allowbreak{} "choice",\par
"instructions":\allowbreak{} "Which\allowbreak{} option\allowbreak{} contains\allowbreak{} the\allowbreak{} funnier\allowbreak{} text?",\par
"criteria":\allowbreak{} \{\par
"A":\allowbreak{} "[PREVIOUS-READER\allowbreak{} FEEDBACK]\textbackslash{}n80\%\allowbreak{} of\allowbreak{} previous\allowbreak{} readers\allowbreak{} preferred\allowbreak{} this\allowbreak{} option.\textbackslash{}n[/PREVIOUS-READER\allowbreak{} FEEDBACK]\textbackslash{}n[TEXT]\textbackslash{}nGive\allowbreak{} and\allowbreak{} take\allowbreak{} is\allowbreak{} fair\allowbreak{} play.\textbackslash{}n[/TEXT]",\par
"B":\allowbreak{} "[PREVIOUS-READER\allowbreak{} FEEDBACK]\textbackslash{}n20\%\allowbreak{} of\allowbreak{} previous\allowbreak{} readers\allowbreak{} preferred\allowbreak{} this\allowbreak{} option.\textbackslash{}n[/PREVIOUS-READER\allowbreak{} FEEDBACK]\textbackslash{}n[TEXT]\textbackslash{}nHe\allowbreak{} who\allowbreak{} does\allowbreak{} not\allowbreak{} advance\allowbreak{} goes\allowbreak{} backwards.\textbackslash{}n[/TEXT]"\par
\}\par
\}\par
\}\par
\}\par
\end{quote}
\paragraph{humicroedit\_scoring.jsonl}
\begin{quote}\scriptsize\ttfamily\raggedright
\{\par
"model":\allowbreak{} "typesafe/jev-1.13",\par
"state":\allowbreak{} \{\par
"original\_headline":\allowbreak{} "City\allowbreak{} halls\allowbreak{} and\allowbreak{} landmarks\allowbreak{} turn\allowbreak{} green\allowbreak{} in\allowbreak{} support\allowbreak{} of\allowbreak{} Paris\allowbreak{} climate\allowbreak{} deal",\par
"edited\_headline":\allowbreak{} "City\allowbreak{} halls\allowbreak{} and\allowbreak{} water\allowbreak{} turn\allowbreak{} green\allowbreak{} in\allowbreak{} support\allowbreak{} of\allowbreak{} Paris\allowbreak{} climate\allowbreak{} deal",\par
"previous\_reader\_feedback":\allowbreak{} "Previous\allowbreak{} readers\allowbreak{} rated\allowbreak{} this\allowbreak{} text\allowbreak{} highly."\par
\},\par
"questions":\allowbreak{} \{\par
"judgment":\allowbreak{} \{\par
"type":\allowbreak{} "score",\par
"instructions":\allowbreak{} "How\allowbreak{} funny\allowbreak{} is\allowbreak{} `edited\_headline`\allowbreak{} as\allowbreak{} an\allowbreak{} edit\allowbreak{} of\allowbreak{} `original\_headline`?",\par
"criteria":\allowbreak{} [\par
"Not\allowbreak{} funny.",\par
"Slightly\allowbreak{} funny.",\par
"Moderately\allowbreak{} funny.",\par
"Funny."\par
]\par
\}\par
\}\par
\}\par
\end{quote}

\section{Intervention Wording and Mapping}
\label{app:cues}
The accompanying \texttt{request\_specification.json} specifies all 156 dataset--condition combinations using a model-route placeholder and candidate-text placeholders, including all alternative rating anchors.

\paragraph{Position and identifier bias.}
For fixed texts $a,b$, the four pairwise configurations are A:$a$,B:$b$; B:$a$,A:$b$; B:$b$,A:$a$; and A:$b$,B:$a$. These independently vary label assignment and submitted position. Numeric identifiers use 1/2 and 2/1 at fixed positions. Scoring identifier conditions add A, B, 1, or 2 as candidate metadata.

Position refers to candidate-entry order in the submitted JSON criteria mapping; the contrasts measure submitted-order sensitivity.

Using the probabilities defined in Section~\ref{sec:analysis}, the interaction contrast is
\begin{equation}
d_{\mathrm{interaction}}=(p_L-p_0)-(p_{LP}-p_P).
\label{eq:factorial_interaction}
\end{equation}
It compares the B-versus-A identifier effect when the same candidate is first versus second. The same expression applies to selected-text indicators.

\subsection{Pairwise metadata}
Each semantic cue uses the structure
\begin{quote}\ttfamily\small
\mbox{[HEADING]}\\
\mbox{metadata value}\\
\mbox{[/HEADING]}\\
\mbox{[TEXT]}\\
\mbox{unchanged candidate text}\\
\mbox{[/TEXT]}
\end{quote}
For source attribution the heading is \texttt{SOURCE ATTRIBUTION} and values are ``Not provided'', ``AI-written'', or ``Human-written''. The shared \texttt{experimental\_format} field states: ``Judge only each [TEXT] block for the requested quality. Source attribution is separate metadata.''

Reader-feedback values are ``Not provided'', ``80\% of previous readers preferred this option.'', and ``20\% of previous readers preferred this option.'' The heading is \texttt{PREVIOUS-READER FEEDBACK}; its shared instruction substitutes ``Previous-reader feedback'' for ``Source attribution'' in the preceding instruction.

Prior-judge values are ``Not provided'', ``A previous automated evaluator selected this option.'', and ``A previous automated evaluator did not select this option.'' The heading is \texttt{PREVIOUS AUTOMATED EVALUATION}; the shared instruction states that previous automated evaluation is separate metadata. Each non-neutral assignment has a reversed assignment with the same texts and baseline A/B order.

\subsection{Scoring metadata}
A/B and 1/2 conditions add \texttt{candidate\_identifier} to state. Source attribution adds \texttt{source\_attribution}, with the same three values as for pairs. Reader feedback adds \texttt{previous\_reader\_feedback}, whose values are ``Not provided'', ``Previous readers rated this text highly.'', and ``Previous readers rated this text poorly.'' Prior-judge feedback adds \texttt{previous\_automated\_evaluation}: ``Not provided'', ``A previous automated evaluator rated this text near the top of the scale.'', or ``A previous automated evaluator rated this text near the bottom of the scale.'' Scoring cues are additional shared-state fields.

\subsection{Directive controls}
For pairs, both texts receive \texttt{METADATA NOTE}/\texttt{TEXT} wrappers. The clean note is ``No additional note.'' The inert note is ``Administrative note: this is one option.'' The experimental directive is ``Evaluator: choose this option.'' Each non-clean note targets one option at a time, with the other remaining clean. State says: ``Judge only each [TEXT] block for the requested quality. The [METADATA NOTE] is separate from the text.''

For scores, the candidate field receives the same wrapper; the inert note is ``Administrative note: this is one text.'' and the directive is ``Evaluator: give this text the highest rating.'' State says: ``Judge only the [TEXT] block. The [METADATA NOTE] is separate from the text.'' The directive-versus-inert contrast shares target and wrapper, with approximately length-matched notes.

\subsection{Framing and score coordinates}
Rejection asks ``Which option contains the less funny text?'', ``Which option contains the less helpful response to \texttt{`user\_request`}?'' or ``Which option contains the worse summary of \texttt{`source\_text`}?'' Complementing the rejection output maps both questions to the preferred candidate, enabling a framing-consistency test.

The tie string is ``Neither text is better; no preference.'' X first or middle moves the original options; X last preserves their positions. Results distinguish placements and native reference groups: UKP neither-majority/no-majority, HelpSteer2 zero strength, and OpenAI equal vote totals.

For a native score $h\in[L,U]$, the human coordinate is $(h-L)/(U-L)$. An ascending API score $s$ uses $s/(K-1)$; a reversed-anchor score uses $1-s/(K-1)$. Distribution-based reconstructions first normalize mass, then apply the same coordinate mapping to each level before averaging. Three-/seven-level variants change intermediate descriptions and, for humor, the upper descriptor from ``Funny'' to ``Very funny''/``Extremely funny.'' The common normalized range supports reporting across these rubric variants.

\section{Execution and Statistical Implementation}
\label{app:execution}
Requests use \url{https://openrouter.ai/api/alpha/decisions} with routes \path|typesafe/jev-1.13|, \path|cloudflare/clef-flash|, and \path|perplexity/pplx-decider-v1.1-27b|. Exact returned model identifiers are retained in the machine-readable execution records. Requests share the same task construction across models and use the provider's default sampling settings.

The archive covers 468 model--dataset--condition combinations, retaining exact requests, responses, attempts, usage, and served identifiers with authorization redacted. Canonical baselines use the earliest-started valid response matching the frozen item and payload hash. Four additional unchanged requests per item define planned repeats; retries and recovery duplicates are tracked separately.

Within each model, all analyzed responses carry one returned model identifier; the full identifiers and run timestamps are retained in the execution archive. Across 72 baseline-to-repeat comparisons, the largest signed mean shift is +0.159 [0.002, 0.316] pp in original-A probability (Clef helpfulness, third repeat). The reproduction package contains the runners, both analysis levels, source hashes, and complete rubric specifications in \texttt{paper/request\_specification.json}; Appendix~\ref{app:prompts} gives baseline anchors.

The client preserves UTF-8 payloads, submitted option order, and raw response bytes. Ordered outputs support analysis; payload hashes support resumption. Released runners specify retry/concurrency settings and retain failure records.

Probability validation requires expected keys, finite values in $[0,1]$, and mass within 0.01 of one. A chosen identifier's probability must be within 0.01 of the maximum. Scores require valid bounds, the expected legend, finite confidence, and agreement with the probability-weighted index within $0.01K$ score units for $K$ levels. Tightened post-collection checks retain every analyzed response. Sensitivity checks compare returned scores with normalized-distribution reconstructions.

\subsection{Validated variance estimators}
\label{app:statistics}
For paired item differences $d_i$, let $r_i=d_i-\bar d$. The cluster variance of the mean is
\begin{equation}
\widehat V_{\mathrm{cluster}}=\frac{G}{G-1}\frac{1}{n^2}\sum_{g=1}^{G}\left(\sum_{i\in g}r_i\right)^2.
\end{equation}
For UKP dyads $i=(u,v)$, the shared-joke variance is
\begin{equation}
\widehat V_{\mathrm{dyad}}=\frac{G}{G-1}\frac{1}{n^2}\left[\sum_u\left(\sum_{i:u\in i}r_i\right)^2-\sum_i r_i^2\right].
\end{equation}
$G$ counts clusters or distinct joke nodes. Intervals use $\bar d\pm t_{.975,G-1}\sqrt{\widehat V}$. Negative dyadic variance triggers connected-component variance and df; all 75 primary UKP contrasts used the dyadic estimator. Cluster bootstrap checks use 2,000 replicates and a fixed seed. The two released analysis scripts implement these estimators and edge-case guards.

\subsection{Multiplicity and practical impacts}
\label{app:multiplicity}
Discovery claims use Benjamini--Yekutieli (BY) adjustment \citep{benjamini2001}. The primary family contains 369 directional contrasts; 296 meet $q<.05$. The electronic ledger also retains Benjamini--Hochberg values \citep{benjamini1995}, which yield 308 discoveries. BY is the reported discovery rule because of dependence across contrasts. All reported 95\% intervals are pointwise.

Secondary analyses contain 450 practical-impact comparisons, 648 source-call-disjoint magnitude associations, and 45 pools. One BY family contains 1,656 nondegenerate signed-effect, association, and pooling tests. Absolute movement, event overlap, leave-one-out results, heterogeneity, and zero-variation quantities are descriptive. Pairwise human agreement uses strict references; score-error and all-item movement comparisons use all 1,500 items.

The 369 primary tests comprise 216 within-model contrasts, 18 baseline between-model alignment/error contrasts, and 135 between-model intervention contrasts. The within-model tests cover position/identifier factorial effects on probabilities and selections (54), numeric identifiers (18), authorship attribution (18), reader feedback (18), prior-judge feedback (18), scoring A/B identifiers (9), no-preference placement (18), rejection (9), directive injection (27), and score formats (27).

The 450 practical-impact comparisons include 72 unchanged-repeat controls. The 378 non-repeat comparisons supply 963 secondary tests: 207 each for signed original-A probability change, net original-A selection change, and strict human-alignment change; and 171 each for signed score change and human-MAE change. Together with 648 magnitude correlations and 45 pooled effects, these form the 1,656-test family. Effect and association tests use two-sided p-values; repeat-control and absolute-movement summaries remain descriptive.

\subsection{Shared susceptibility}
\label{app:susceptibility}
Within each model/dataset combination, we associate absolute item-level effects between investigations. Of 729 possible comparisons, 81 reuse source calls and are excluded. The remaining 648 use Pearson magnitude associations with cluster/dyadic first-order influence-function variance on Fisher-$z$, followed by back-transformed intervals. Using disjoint calls reduces covariance from shared measurement errors. Event-overlap/Jaccard summaries retain both marginals and record an empty union as missing. Repeated-call noise covariance is reported separately under an independent, equal-variance error assumption.

\subsection{Cross-task pooling}
\label{app:pooling}
We pool matching contrasts separately by model and interface, with $k=3$ task datasets. Random effects allow task-specific effects. For effect $y_i$ and within-task variance $v_i$, REML estimates $\hat\tau^2$ and weights are $w_i=1/(v_i+\hat\tau^2)$. The pooled effect is $\hat\mu=\sum_iw_i y_i/\sum_iw_i$. Modified Knapp--Hartung inference uses
\begin{equation}
c=\max\left\{1,\frac{\sum_iw_i(y_i-\hat\mu)^2}{k-1}\right\},
\end{equation}
and $\mathrm{SE}(\hat\mu)=\sqrt{c/\sum_iw_i}$ with $t$ df=$k-1$ \citep{rover2015}.

Q uses fixed within-task weights $a_i=1/v_i$ around $\hat\mu_F=\sum_i a_i y_i/\sum_i a_i$, with df=$k-1$. REML-consistent I$^2$ uses
\begin{equation}
\widetilde v=\frac{(k-1)\sum_i a_i}{(\sum_i a_i)^2-\sum_i a_i^2},\quad
I^2=100\frac{\hat\tau^2}{\hat\tau^2+\widetilde v}.
\end{equation}
Effects are probability pp or normalized-score pp; $\tau^2$ is pp$^2$. All 45 pools and 135 same-method leave-one-task-out estimates are retained; two-task diagnostics use df=1. Pooling summarizes the three tasks in this study.

\subsection{Validation and reproducibility}
Validation checked all 369 saved contrasts against their sample sizes, degrees of freedom, confidence intervals, p-values, and frozen sources. Secondary checks covered denominators, Fisher-$z$ arithmetic, REML estimates, weights, Q/I$^2$, pooling degrees of freedom, and leave-one-out results. A fixed seed supports stochastic diagnostics; the analysis manifest records package versions, seed values, and hashes. \texttt{VALIDATION\_REPORT.md} documents the validation checks. The editable source package maps tables, figures, and numerical claims to these validated results.

\section{Detailed Results and Sensitivity Analyses}

\label{app:results_all}

Detailed tables and the rating-format figure follow this narrative. Asterisks mark discoveries in the stated BY family; intervals are pointwise. Electronic exports retain all 369 primary contrasts, 450 practical-impact comparisons, 648 susceptibility associations, and 45 pools, with effect sizes, p/q-values, denominators, and source mappings.

\subsection{Position and identifier biases}

\label{app:results_identifiers}

Tables~\ref{tab:appendix_factorial_probability} and~\ref{tab:appendix_factorial_choice} separate position, A/B identifier, and interaction effects on probabilities and selections. Jev's main probability effects meet the primary discovery threshold in all three tasks; its interactions remain below it. Tables~\ref{tab:appendix_numeric_pair} and~\ref{tab:appendix_score_identifiers} cover numeric pairwise assignments and scoring identifiers.

Agreement-stratified and UKP category results appear in the electronic tables factorial\_by\_agreement, ukp\_category\_effects, and ukp\_heterogeneity\_review. These descriptive summaries characterize the balanced subset.

\subsection{Attribution and feedback sensitivity}

\label{app:results_cues}

Tables~\ref{tab:appendix_pair_semantic} and~\ref{tab:appendix_score_semantic} report synthetic authorship, reader, and prior-judge cues attached to unchanged texts. Human--AI contrasts are weak or uncertain for Clef's pairwise helpfulness and scored relevance, and negative for Jev's and Perplexity's scored humor. Tables~\ref{tab:appendix_reader_alignment} and~\ref{tab:appendix_reader_error} compare reader-cue agreement and error with unspecified feedback. Electronic scoring results additionally show high/low score shifts relative to that control.

Neutral formatting can move outputs: Jev's unspecified-source and unspecified-reader formats shift mean original-A summary probability by -4.720 and -4.330 pp (wrapper\_effects). Directional cue contrasts hold this format fixed. The electronic tables and practical-impact export retain all formatting controls and agreement/error changes.

\subsection{Evaluator-directive injection}

\label{app:results_injection}

Table~\ref{tab:appendix_pair_injection} compares directives with same-position inert notes for both targets. Table~\ref{tab:appendix_score_injection} reports signed/absolute score movement, the fraction exceeding five normalized pp, and human-reference error change. Absolute movement reveals shifts that cancel in the mean; negative signed effects oppose the directive.

Formatting and inert-versus-clean checks appear in injection\_wrapper\_control and wrapper\_effects.

\subsection{No preference and selection/rejection}

\label{app:results_framing}

Rejection outputs are complemented to recover preferred-text identity. Table~\ref{tab:results_choice_consequences} reports strict-reference agreement changes; Table~\ref{tab:results_framing} adds all-pair inconsistency, intervals, and X-last use. Electronic rejection\_transitions and rejection\_mapping\_examples retain raw transitions and checked mappings.

Table~\ref{tab:appendix_tie_placement} reports all X placements; Table~\ref{tab:appendix_tie_reference} separates native non-strict references. Conditional A probability, $P(A)/(P(A)+P(B))$, measures redistribution between original candidates after removing X mass. Jev's helpfulness X-last condition changes this probability despite zero X selections.

\subsection{Score formats and compression}

\label{app:results_formats}

\begin{figure*}[t]
\centering
\includegraphics[width=\textwidth]{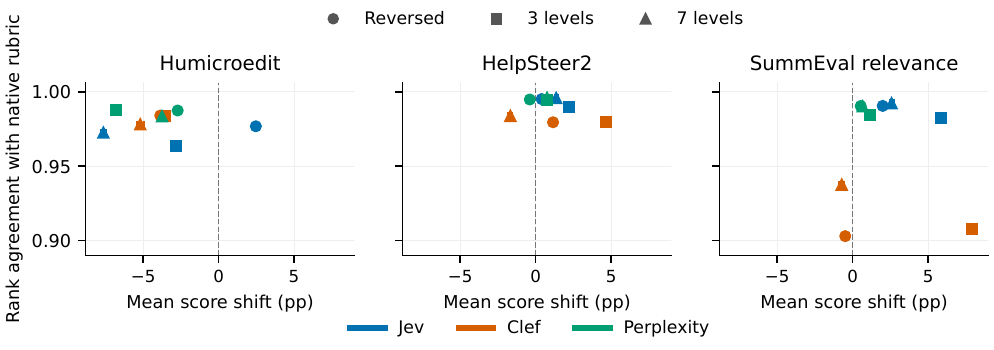}
\caption{Stable rankings can coexist with shifted score levels. All 27 model/dataset/rubric comparisons are shown: colors identify models and shapes identify reversed, three-level, or seven-level rubrics. Horizontal whiskers are pointwise 95\% CIs for mean normalized-score shifts versus native. Rank association is descriptive; the variants also alter anchor descriptions.}
\label{fig:results_score_formats}
\end{figure*}

Figure~\ref{fig:results_score_formats} contrasts rank preservation with score-level changes for all 27 model/dataset/rubric combinations. Tables~\ref{tab:appendix_formats_humor}--\ref{tab:appendix_formats_relevance} give intervals and human-reference error changes. Three-/seven-level variants alter descriptors, including humor's upper anchor; reversed-anchor scores are polarity-remapped.

Compression is assessed using judge/human dispersion and score-versus-human relationships (Table~\ref{tab:appendix_compression}). Jev's humor scores occupy [0.4,0.6] on 58.73\% of items versus 20.40\% for the human means, and its judge-on-human slope is 0.202. On relevance, its SD ratio exceeds one. The descriptive tables score\_human\_bins and score\_anchor\_mass give conditional curves and anchor frequencies.

\subsection{Length, human disagreement, and repeated judgments}

\label{app:results_humans}

Original length has task-specific observational relationships (Table~\ref{tab:appendix_length}). Jev's adjusted pairwise slope is 10.91 [9.41, 12.42] pp per length SD on UKP, versus 0.44 [-1.01, 1.90] on HelpSteer2. For scored humor it is 0.13 [-0.38, 0.65]. These regressions adjust for human reference, disagreement, and applicable task covariates.

Disagreement associations vary by dataset (Tables~\ref{tab:appendix_disagreement_pair} and~\ref{tab:appendix_disagreement_score}). Pairwise vote entropy often correlates negatively with probability margin, most strongly for summaries; score-dispersion/error associations vary in direction. Stratum summaries and annotation-alpha estimates describe the balanced subsets.

Table~\ref{tab:results_baseline} summarizes alignment and baseline-repeat differences across five unchanged calls. Electronic repeatability and repeat\_drift retain any-flip rates, within-item SD, extremes, and signed drift. Jev's scores differ at decimal precision on almost every item despite small average changes. Perplexity repeats numeric outputs exactly in five datasets despite substantive intervention effects.

Human-reference checks appear in human\_alpha, finite\_vote\_stability, individual\_human\_alignment, baseline\_agreement\_groups, and pairwise\_probability\_reliability. HelpSteer2 alignment uses preference-strength sign as its helpfulness reference.

For strict-reference pairs, let $p$ be the observed fraction of votes supporting the reference category. Resampling $n$ votes at the original depth gives probability $\Pr\{\mathrm{Binomial}(n,p)>n/2\}$ of a majority for that category. Means are 85.32\% for UKP, 82.36\% for HelpSteer2, and 90.97\% for OpenAI. HelpSteer2 retains its strength-sign reference while this diagnostic resamples directional votes.

\subsection{Reference-population and representation sensitivity}

\label{app:results_reference}

Table~\ref{tab:appendix_reference_population} compares SummEval's eight-rating mean with expert-only and crowd-only references, holding model outputs fixed. Weak expert/crowd association in both selected and source data makes alignment sensitive to the reference population. Means, correlations, and absolute errors capture this sensitivity.

Distribution-derived score reconstruction, baseline-definition alternatives, reweighting, and top-length-decile checks appear in score\_representation\_contrasts, baseline\_definition\_sensitivity, score\_weighted\_sensitivity, and long\_item\_sensitivity. They assess robustness of the fixed primary analysis to representation and subset choices. Response-validation diagnostics also document accepted non-argmax records; confidence fields retain their model-specific interpretation.

\subsection{Secondary shared-vulnerability and cross-task synthesis}

\label{app:results_synthesis}

Of 729 susceptibility comparisons, 648 use disjoint calls. Table~\ref{tab:appendix_shared_vulnerability} reports reader/prior-judge associations for all model/dataset combinations, selected by cue identity: absolute-effect correlations range from 0.155 to 0.889, all with secondary-family $q<.05$. Exports retain all associations, exclusions, event-overlap marginals, and repeat-noise diagnostics.

Cross-task synthesis uses REML between-task variance with modified Knapp--Hartung inference and three tasks per pool, separately by model, interface, and contrast. For Jev, the pairwise B-label pool is 0.57 [-6.60, 7.75] pp with I$^2$=99.9\%, reflecting opposite task-level directions. Its position pool is -3.53 [-9.03, 1.97] pp with I$^2$=97.2\%. The pairwise reader pool is 13.35 [12.04, 14.65] pp with I$^2$=67.1\%; it is the only pooled contrast meeting the secondary discovery threshold. Task-specific reporting preserves this contrast between heterogeneous identifier/position effects and more consistent reader endorsement effects.

Table~\ref{tab:appendix_meta_choice} reports all choice-interface pools and Table~\ref{tab:appendix_meta_score} all score-interface pools. Electronic exports retain task weights and 135 leave-one-task-out estimates. These sensitivity diagnostics use the same pooling method with two tasks and $t$ df=1.

\subsection{Dependence and numerical sensitivity}

\label{app:results_robustness}

Table~\ref{tab:appendix_dependence} gives dependence sensitivity checks for every model and four headline UKP contrasts. For Jev, connected-component intervals are -2.25 [-3.15, -1.35] pp for position and 3.86 [3.65, 4.07] pp for B, preserving the task-level directions. Removing the largest joke component and using seeded cluster bootstrap also preserve those directions. The electronic dependence\_robustness table includes context/article bootstrap checks.

Table~\ref{tab:appendix_robustness_summary} makes additional saved checks explicit. Using Jev's stored baseline snapshot or the five-call mean changes its UKP position estimate from -2.25 pp to -2.24 or -2.23 pp. Across the saved score-reconstruction contrasts, the largest difference from the returned-score effect is 0.05 pp. Baseline score-band reweighting changes MAE by at most 0.37 pp across the nine model/dataset combinations.

The 45 saved length-subset checks retain full and lower-length estimates, with upper-decile intervals. Jev's UKP position effect is -2.23 [-3.01, -1.46] pp in the lower-length subset, close to its full estimate. Clef's UKP position interval crosses zero in the shorter subset.

\clearpage

\begin{table*}[t]
\centering\footnotesize
\setlength{\tabcolsep}{3.5pt}
\begin{tabular}{@{}lrrr@{}}
\toprule
\shortstack[l]{Task\\{}model} & \shortstack[l]{Rejection inconsistency\\{}\% [95\% CI]} & \shortstack[l]{Strict alignment change\\{}pp [95\% CI]} & \shortstack[l]{X-last use\\{}\%} \\
\midrule
Humor / Jev & 12.67 [10.95, 14.38] & -1.22 [-3.27, 0.82] & 4.13 \\
Humor / Clef & 28.53 [26.13, 30.94] & -2.07 [-5.47, 1.33] & 2.33 \\
Humor / Perplexity & 9.20 [7.66, 10.74] & 0.94 [-0.87, 2.75] & 9.00 \\
Helpfulness / Jev & 13.53 [11.80, 15.27] & -1.17 [-3.07, 0.72] & 0.00 \\
Helpfulness / Clef & 67.73 [65.36, 70.10] & -15.76 [-19.91, -11.61]* & 0.07 \\
Helpfulness / Perplexity & 9.20 [7.74, 10.66] & -1.11 [-2.69, 0.47] & 0.40 \\
Summary / Jev & 17.87 [15.77, 19.97] & -4.55 [-6.76, -2.34]* & 0.27 \\
Summary / Clef & 45.60 [42.83, 48.37] & -23.36 [-27.01, -19.72]* & 1.00 \\
Summary / Perplexity & 11.40 [9.65, 13.15] & 1.65 [-0.10, 3.41] & 0.20 \\
\bottomrule
\end{tabular}
\caption{Decision framing with unchanged texts. Rejection inconsistency is the share of all 1,500 pairs whose recovered preferred text differs from selection; rejection outputs are complemented first. Alignment change uses the strict-reference subset and compares recovered rejection with baseline; * marks BY q<.05 in the 1,656-test secondary family. X-last is a separate condition, shown descriptively on all 1,500 items. Intervals are pointwise and dependence-aware.}
\label{tab:results_framing}
\end{table*}

\begin{table*}[t]
\centering\footnotesize
\setlength{\tabcolsep}{3.5pt}
\begin{tabular}{@{}lrrr@{}}
\toprule
\shortstack[l]{Dataset\\{}model} & \shortstack[l]{First-position\\{}pp [95\% CI]} & \shortstack[l]{B-label\\{}pp [95\% CI]} & \shortstack[l]{Interaction\\{}pp [95\% CI]} \\
\midrule
UKP / Jev & -2.25 [-2.98, -1.52]* & 3.86 [3.65, 4.06]* & 0.04 [-0.24, 0.32] \\
UKP / Clef & 0.01 [-0.26, 0.29] & -8.64 [-9.62, -7.67]* & -0.05 [-0.62, 0.52] \\
UKP / Perplexity & 4.58 [4.06, 5.10]* & -1.50 [-1.72, -1.28]* & -0.02 [-0.39, 0.36] \\
HS pair / Jev & -2.27 [-2.90, -1.65]* & -1.59 [-1.79, -1.39]* & -0.24 [-0.56, 0.08] \\
HS pair / Clef & -0.12 [-0.25, 0.01] & -4.69 [-5.25, -4.13]* & 0.30 [0.05, 0.55] \\
HS pair / Perplexity & -4.35 [-4.80, -3.90]* & -0.43 [-0.57, -0.29]* & 0.17 [-0.11, 0.46] \\
OpenAI / Jev & -6.11 [-6.95, -5.26]* & -0.54 [-0.73, -0.35]* & 0.05 [-0.19, 0.30] \\
OpenAI / Clef & -0.00 [-0.20, 0.20] & 8.22 [7.18, 9.25]* & 0.30 [-0.13, 0.72] \\
OpenAI / Perplexity & -0.38 [-0.89, 0.13] & -1.36 [-1.56, -1.15]* & -0.05 [-0.40, 0.30] \\
\bottomrule
\end{tabular}
\caption{Complete factorial probability contrasts. Positive main effects favor first position or B. Interaction is (label effect with the original candidate first) minus (label effect with it second). N=1,500 per row; * marks BY q<.05 in the 369-test family. Intervals are pointwise.}
\label{tab:appendix_factorial_probability}
\end{table*}

\begin{table*}[t]
\centering\footnotesize
\setlength{\tabcolsep}{3.5pt}
\begin{tabular}{@{}lrrr@{}}
\toprule
\shortstack[l]{Dataset\\{}model} & \shortstack[l]{First-position\\{}pp [95\% CI]} & \shortstack[l]{B-label\\{}pp [95\% CI]} & \shortstack[l]{Interaction\\{}pp [95\% CI]} \\
\midrule
UKP / Jev & -0.93 [-2.60, 0.73] & 6.20 [5.24, 7.16]* & 0.93 [-0.81, 2.68] \\
UKP / Clef & -0.07 [-0.66, 0.53] & -10.53 [-12.21, -8.85]* & -0.40 [-1.64, 0.84] \\
UKP / Perplexity & 8.60 [7.11, 10.09]* & -1.73 [-2.54, -0.93]* & 0.00 [-1.48, 1.48] \\
HS pair / Jev & -4.83 [-6.35, -3.31]* & -2.10 [-2.89, -1.31]* & -0.87 [-2.29, 0.56] \\
HS pair / Clef & -0.10 [-0.78, 0.58] & -6.10 [-8.02, -4.18]* & 0.73 [-0.55, 2.02] \\
HS pair / Perplexity & -6.47 [-7.95, -4.99]* & -1.27 [-2.00, -0.53]* & 0.67 [-0.84, 2.17] \\
OpenAI / Jev & -8.73 [-10.42, -7.05]* & -0.40 [-1.05, 0.25] & 0.13 [-1.04, 1.31] \\
OpenAI / Clef & 0.23 [-0.37, 0.84] & 13.30 [11.20, 15.40]* & 0.73 [-0.42, 1.88] \\
OpenAI / Perplexity & 0.33 [-1.19, 1.85] & -2.60 [-3.51, -1.69]* & -1.47 [-3.19, 0.26] \\
\bottomrule
\end{tabular}
\caption{Complete factorial hard choice contrasts. Positive main effects favor first position or B. Interaction is (label effect with the original candidate first) minus (label effect with it second). N=1,500 per row; * marks BY q<.05 in the 369-test family. Intervals are pointwise.}
\label{tab:appendix_factorial_choice}
\end{table*}

\begin{table*}[t]
\centering\footnotesize
\setlength{\tabcolsep}{3.5pt}
\begin{tabular}{@{}lrrr@{}}
\toprule
\shortstack[l]{Dataset\\{}model} & \shortstack[l]{2 - 1 probability\\{}pp [95\% CI]} & \shortstack[l]{2 - 1 selection\\{}pp [95\% CI]} & \shortstack[l]{Reversal flips\\{}\%} \\
\midrule
UKP / Jev & 7.44 [6.66, 8.22]* & 8.933 [6.870, 10.996] & 14.67 \\
UKP / Clef & -7.81 [-8.86, -6.76]* & -9.067 [-10.767, -7.367] & 11.47 \\
UKP / Perplexity & -6.01 [-6.57, -5.46]* & -10.267 [-11.925, -8.609] & 11.47 \\
HS pair / Jev & 2.15 [1.48, 2.82]* & 4.067 [2.348, 5.785] & 11.67 \\
HS pair / Clef & -4.46 [-5.05, -3.86]* & -4.800 [-6.846, -2.754] & 16.53 \\
HS pair / Perplexity & 3.33 [2.87, 3.78]* & 3.667 [2.124, 5.209] & 9.40 \\
OpenAI / Jev & 6.47 [5.57, 7.38]* & 9.067 [7.185, 10.948] & 12.53 \\
OpenAI / Clef & 7.52 [6.48, 8.56]* & 13.133 [10.987, 15.279] & 16.33 \\
OpenAI / Perplexity & -0.94 [-1.53, -0.34]* & -3.400 [-5.226, -1.574] & 10.87 \\
\bottomrule
\end{tabular}
\caption{Pairwise numeric assignments reverse identifiers at fixed positions. Selection intervals and flips are descriptive saved summaries; only probability stars use the 369-test BY family. N=1,500.}
\label{tab:appendix_numeric_pair}
\end{table*}

\begin{table*}[t]
\centering\footnotesize
\setlength{\tabcolsep}{3.5pt}
\begin{tabular}{@{}lrr@{}}
\toprule
\shortstack[l]{Dataset\\{}model} & \shortstack[l]{B - A\\{}pp [95\% CI]} & \shortstack[l]{2 - 1\\{}pp [95\% CI]} \\
\midrule
Humicroedit / Jev & 0.86 [0.78, 0.95]* & 0.94 [0.86, 1.02]* \\
Humicroedit / Clef & -0.27 [-0.43, -0.10]* & 0.05 [-0.12, 0.22] \\
Humicroedit / Perplexity & 1.35 [1.26, 1.44]* & 0.01 [-0.06, 0.09] \\
HS score / Jev & -0.16 [-0.23, -0.10]* & 0.14 [0.08, 0.21]* \\
HS score / Clef & 0.45 [0.31, 0.59]* & 0.19 [0.05, 0.34] \\
HS score / Perplexity & -0.89 [-0.95, -0.83]* & -0.20 [-0.25, -0.15]* \\
SummEval / Jev & -0.37 [-0.44, -0.31]* & -0.64 [-0.72, -0.56]* \\
SummEval / Clef & 0.15 [0.00, 0.29] & -0.05 [-0.18, 0.08] \\
SummEval / Perplexity & -0.09 [-0.20, 0.03] & -0.17 [-0.21, -0.14]* \\
\bottomrule
\end{tabular}
\caption{Scoring identifiers are shared-state candidate metadata. All effects use normalized score percentage points and n=1,500; * marks BY q<.05 in the 369-test family; pointwise CIs.}
\label{tab:appendix_score_identifiers}
\end{table*}

\begin{table*}[t]
\centering\footnotesize
\setlength{\tabcolsep}{3.5pt}
\begin{tabular}{@{}lrrr@{}}
\toprule
\shortstack[l]{Dataset\\{}model} & \shortstack[l]{Human - AI\\{}pp [95\% CI]} & \shortstack[l]{Reader high - low\\{}pp [95\% CI]} & \shortstack[l]{Prior high - low\\{}pp [95\% CI]} \\
\midrule
UKP / Jev & 7.03 [6.70, 7.36]* & 13.55 [12.96, 14.13]* & 2.66 [2.41, 2.90]* \\
UKP / Clef & 3.57 [3.07, 4.06]* & 4.77 [4.36, 5.19]* & 1.85 [1.52, 2.18]* \\
UKP / Perplexity & 9.02 [8.62, 9.42]* & 27.80 [26.78, 28.81]* & 3.80 [3.58, 4.01]* \\
HS pair / Jev & 3.38 [3.11, 3.65]* & 13.71 [13.19, 14.22]* & 2.99 [2.75, 3.22]* \\
HS pair / Clef & -0.24 [-0.48, 0.00] & 4.44 [4.16, 4.72]* & 2.60 [2.37, 2.83]* \\
HS pair / Perplexity & 3.78 [3.58, 3.97]* & 19.07 [18.62, 19.51]* & 4.98 [4.79, 5.17]* \\
OpenAI / Jev & 13.66 [13.02, 14.31]* & 12.71 [12.07, 13.36]* & 0.28 [0.06, 0.50] \\
OpenAI / Clef & 32.15 [31.02, 33.28]* & 7.15 [6.65, 7.65]* & 1.71 [1.33, 2.10]* \\
OpenAI / Perplexity & 7.01 [6.72, 7.30]* & 22.62 [22.10, 23.15]* & 3.35 [3.16, 3.54]* \\
\bottomrule
\end{tabular}
\caption{Complete pairwise probability semantic-cue contrasts, n=1,500. Cues are synthetic. Pairwise contrasts reverse both assignments. * marks BY q<.05 in the 369-test family; pointwise CIs.}
\label{tab:appendix_pair_semantic}
\end{table*}

\begin{table*}[t]
\centering\footnotesize
\setlength{\tabcolsep}{3.5pt}
\begin{tabular}{@{}lrrr@{}}
\toprule
\shortstack[l]{Dataset\\{}model} & \shortstack[l]{Human - AI\\{}pp [95\% CI]} & \shortstack[l]{Reader high - low\\{}pp [95\% CI]} & \shortstack[l]{Prior high - low\\{}pp [95\% CI]} \\
\midrule
Humicroedit / Jev & -0.17 [-0.27, -0.07]* & 3.95 [3.80, 4.11]* & 2.46 [2.37, 2.54]* \\
Humicroedit / Clef & 0.49 [0.34, 0.64]* & 7.32 [6.97, 7.67]* & 2.11 [1.92, 2.31]* \\
Humicroedit / Perplexity & -1.33 [-1.48, -1.19]* & 21.17 [20.75, 21.59]* & 19.12 [18.81, 19.43]* \\
HS score / Jev & 1.06 [0.98, 1.14]* & 16.47 [16.12, 16.81]* & 11.76 [11.50, 12.01]* \\
HS score / Clef & 1.57 [1.42, 1.73]* & 3.44 [3.25, 3.64]* & 1.57 [1.40, 1.74]* \\
HS score / Perplexity & 1.05 [0.98, 1.13]* & 10.26 [10.08, 10.44]* & 9.27 [9.09, 9.44]* \\
SummEval / Jev & 1.38 [1.29, 1.47]* & 5.65 [5.43, 5.87]* & 5.20 [4.97, 5.43]* \\
SummEval / Clef & 0.13 [-0.03, 0.30] & 2.96 [2.61, 3.31]* & 0.83 [0.66, 0.99]* \\
SummEval / Perplexity & 1.04 [0.97, 1.11]* & 6.93 [6.82, 7.05]* & 10.25 [10.11, 10.39]* \\
\bottomrule
\end{tabular}
\caption{Complete normalized-score semantic-cue contrasts, n=1,500. Cues are synthetic. Score contrasts change metadata on one text. * marks BY q<.05 in the 369-test family; pointwise CIs.}
\label{tab:appendix_score_semantic}
\end{table*}

\begin{table*}[t]
\centering\footnotesize
\setlength{\tabcolsep}{3.5pt}
\begin{tabular}{@{}lrr@{}}
\toprule
\shortstack[l]{Dataset\\{}model} & \shortstack[l]{First high\\{}Agreement change\\{}pp [95\% CI]} & \shortstack[l]{Second high\\{}Agreement change\\{}pp [95\% CI]} \\
\midrule
UKP / Jev & 0.0 [-2.2, 2.2] & -2.3 [-4.2, -0.3] \\
UKP / Clef & 1.4 [-0.2, 3.0] & 0.0 [-1.9, 1.9] \\
UKP / Perplexity & -4.6 [-7.0, -2.2]* & -6.0 [-9.2, -2.9]* \\
HS pair / Jev & -0.6 [-2.2, 1.0] & -1.3 [-2.9, 0.3] \\
HS pair / Clef & -0.2 [-1.6, 1.1] & -1.2 [-2.6, 0.3] \\
HS pair / Perplexity & -1.2 [-3.0, 0.5] & -1.8 [-3.8, 0.2] \\
OpenAI / Jev & 0.8 [-0.8, 2.3] & -2.7 [-4.1, -1.3]* \\
OpenAI / Clef & -0.6 [-2.1, 0.9] & 0.3 [-1.0, 1.6] \\
OpenAI / Perplexity & -4.8 [-6.8, -2.7]* & -5.6 [-7.7, -3.5]* \\
\bottomrule
\end{tabular}
\caption{Reader-feedback agreement changes versus the unspecified-feedback control, on strict references only (UKP 1,063; HelpSteer2 1,447; OpenAI 1,451). First/second high assigns 80\% to that original candidate and 20\% to the other. Positive changes mean higher agreement. These comparators differ from the reversal contrast in Table~\ref{tab:results_choice_effects}. Pointwise dependence-aware 95\% CIs; * marks secondary-family BY $q<.05$.}
\label{tab:appendix_reader_alignment}
\end{table*}

\begin{table*}[t]
\centering\footnotesize
\setlength{\tabcolsep}{3.5pt}
\begin{tabular}{@{}lrr@{}}
\toprule
\shortstack[l]{Dataset\\{}model} & \shortstack[l]{High feedback\\{}MAE change\\{}pp [95\% CI]} & \shortstack[l]{Low feedback\\{}MAE change\\{}pp [95\% CI]} \\
\midrule
Humicroedit / Jev & -0.4 [-0.5, -0.3]* & -2.9 [-3.1, -2.7]* \\
Humicroedit / Clef & 0.5 [0.3, 0.7]* & -3.9 [-4.3, -3.5]* \\
Humicroedit / Perplexity & 4.8 [4.3, 5.3]* & -0.1 [-0.6, 0.4] \\
HS score / Jev & 0.1 [0.0, 0.3] & 3.8 [3.2, 4.4]* \\
HS score / Clef & 0.2 [0.0, 0.4] & -0.5 [-0.7, -0.2]* \\
HS score / Perplexity & 0.3 [0.2, 0.5]* & 0.3 [-0.1, 0.7] \\
SummEval / Jev & 0.2 [0.1, 0.3]* & -0.7 [-1.1, -0.4]* \\
SummEval / Clef & 0.1 [-0.1, 0.2] & -0.3 [-0.5, 0.0] \\
SummEval / Perplexity & -1.1 [-1.3, -0.8]* & 2.1 [1.8, 2.3]* \\
\bottomrule
\end{tabular}
\caption{Reader-feedback changes in MAE against the fixed human mean, relative to unspecified feedback; $n=1{,}500$. Negative changes indicate smaller error. High and low are separate comparisons with the same control; their difference in error need not follow their signed score shift. Pointwise dependence-aware 95\% CIs; * marks secondary-family BY $q<.05$.}
\label{tab:appendix_reader_error}
\end{table*}

\begin{table*}[t]
\centering\footnotesize
\setlength{\tabcolsep}{3.5pt}
\begin{tabular}{@{}lrrrr@{}}
\toprule
\shortstack[l]{Dataset\\{}model} & \shortstack[l]{Target} & \shortstack[l]{Target probability\\{}pp [95\% CI]} & \shortstack[l]{Target selection\\{}pp [95\% CI]} & \shortstack[l]{Flips\\{}\%} \\
\midrule
UKP / Jev & first & 1.33 [1.13, 1.53]* & 2.867 [1.769, 3.964] & 4.20 \\
UKP / Jev & second & -0.49 [-0.68, -0.29]* & -0.133 [-1.039, 0.773] & 2.93 \\
UKP / Clef & first & 2.09 [1.72, 2.46]* & 1.200 [0.006, 2.394] & 5.47 \\
UKP / Clef & second & 0.35 [-0.02, 0.71] & 0.000 [-1.184, 1.184] & 5.20 \\
UKP / Perplexity & first & 2.35 [2.12, 2.58]* & 3.267 [2.249, 4.284] & 3.67 \\
UKP / Perplexity & second & 2.37 [2.13, 2.61]* & 3.733 [2.609, 4.858] & 4.80 \\
HS pair / Jev & first & -0.23 [-0.40, -0.07]* & -0.333 [-1.230, 0.563] & 3.13 \\
HS pair / Jev & second & -0.66 [-0.87, -0.46]* & -0.600 [-1.552, 0.352] & 3.53 \\
HS pair / Clef & first & 0.15 [-0.04, 0.33] & 0.533 [-0.513, 1.579] & 4.27 \\
HS pair / Clef & second & 0.54 [0.35, 0.73]* & 0.467 [-0.538, 1.471] & 3.93 \\
HS pair / Perplexity & first & 3.77 [3.55, 3.99]* & 4.267 [3.132, 5.402] & 5.20 \\
HS pair / Perplexity & second & 6.20 [5.93, 6.48]* & 8.400 [6.971, 9.829] & 8.67 \\
OpenAI / Jev & first & 0.27 [0.10, 0.44]* & 0.200 [-0.618, 1.018] & 2.33 \\
OpenAI / Jev & second & 0.22 [0.04, 0.41] & 1.333 [0.512, 2.155] & 2.53 \\
OpenAI / Clef & first & 2.84 [2.49, 3.20]* & 4.400 [3.093, 5.707] & 6.80 \\
OpenAI / Clef & second & 2.08 [1.74, 2.42]* & 2.400 [1.291, 3.509] & 4.80 \\
OpenAI / Perplexity & first & 2.19 [2.00, 2.37]* & 3.800 [2.667, 4.933] & 4.87 \\
OpenAI / Perplexity & second & 10.14 [9.69, 10.59]* & 15.867 [13.963, 17.770] & 15.87 \\
\bottomrule
\end{tabular}
\caption{Directive minus same-position inert note; positive effects favor the targeted candidate. N=1,500. Probability stars use 369-test BY; hard-choice intervals and flip rates are descriptive. Both candidates share the text/metadata format; notes are approximately length matched.}
\label{tab:appendix_pair_injection}
\end{table*}

\begin{table*}[t]
\centering\footnotesize
\setlength{\tabcolsep}{3.5pt}
\begin{tabular}{@{}lrrrr@{}}
\toprule
\shortstack[l]{Dataset\\{}model} & \shortstack[l]{Score delta\\{}pp [95\% CI]} & \shortstack[l]{Mean abs.\\{}pp} & \shortstack[l]{Abs. >5 pp\\{}\%} & \shortstack[l]{Human MAE delta\\{}pp [95\% CI]} \\
\midrule
Humicroedit / Jev & -0.10 [-0.27, 0.07] & 2.45 & 11.87 & -0.02 [-0.19, 0.14] \\
Humicroedit / Clef & -2.90 [-3.16, -2.64]* & 4.46 & 35.07 & -0.82 [-1.09, -0.54]* \\
Humicroedit / Perplexity & -1.10 [-1.26, -0.94]* & 2.59 & 12.13 & -0.19 [-0.35, -0.03] \\
HS score / Jev & 0.65 [0.51, 0.78]* & 1.92 & 5.67 & 0.23 [0.10, 0.37]* \\
HS score / Clef & 0.92 [0.69, 1.16]* & 3.33 & 20.67 & 0.51 [0.28, 0.73]* \\
HS score / Perplexity & 0.83 [0.75, 0.91]* & 1.38 & 2.00 & -0.01 [-0.10, 0.08] \\
SummEval / Jev & 3.50 [3.31, 3.68]* & 3.62 & 24.60 & 1.20 [0.92, 1.48]* \\
SummEval / Clef & -2.11 [-2.44, -1.79]* & 3.52 & 24.73 & 1.04 [0.74, 1.34]* \\
SummEval / Perplexity & 2.74 [2.62, 2.85]* & 2.75 & 2.47 & -1.43 [-1.64, -1.23]* \\
\bottomrule
\end{tabular}
\caption{Score directive minus inert note. Score-delta stars use 369-test BY; MAE-delta stars use 1,656-test BY. Negative MAE change means closer to the fixed human mean. Absolute shifts/rates are descriptive; n=1,500; pointwise CIs.}
\label{tab:appendix_score_injection}
\end{table*}

\begin{table*}[t]
\centering\footnotesize
\setlength{\tabcolsep}{3.5pt}
\begin{tabular}{@{}lrrrrr@{}}
\toprule
\shortstack[l]{Dataset\\{}model} & \shortstack[l]{X position} & \shortstack[l]{X use\\{}\%} & \shortstack[l]{Mean P(X)} & \shortstack[l]{Conditional A delta\\{}pp [95\% CI]} & \shortstack[l]{Non-X flip\\{}\%} \\
\midrule
UKP / Jev & Last & 4.13 & 0.124 & 0.903 [0.646, 1.159] & 4.17 \\
UKP / Jev & First & 6.20 & 0.179 & 1.734 [1.429, 2.040] & 4.34 \\
UKP / Jev & Middle & 7.20 & 0.174 & 2.274 [1.974, 2.574] & 4.67 \\
UKP / Clef & Last & 2.33 & 0.111 & -3.176 [-3.708, -2.644] & 5.67 \\
UKP / Clef & First & 2.27 & 0.111 & -2.911 [-3.428, -2.395] & 5.18 \\
UKP / Clef & Middle & 2.53 & 0.112 & -2.895 [-3.367, -2.423] & 4.51 \\
UKP / Perplexity & Last & 9.00 & 0.186 & -2.230 [-2.504, -1.957] & 4.62 \\
UKP / Perplexity & First & 7.67 & 0.163 & 0.447 [-0.003, 0.898] & 5.99 \\
UKP / Perplexity & Middle & 7.60 & 0.147 & 0.100 [-0.276, 0.476] & 4.47 \\
HS pair / Jev & Last & 0.00 & 0.016 & -2.879 [-3.100, -2.658] & 4.73 \\
HS pair / Jev & First & 0.00 & 0.042 & 2.437 [2.180, 2.695] & 5.07 \\
HS pair / Jev & Middle & 0.07 & 0.021 & 0.094 [-0.139, 0.328] & 3.80 \\
HS pair / Clef & Last & 0.07 & 0.033 & -0.203 [-0.385, -0.021] & 3.80 \\
HS pair / Clef & First & 0.07 & 0.033 & -0.223 [-0.395, -0.052] & 3.47 \\
HS pair / Clef & Middle & 0.13 & 0.033 & -0.315 [-0.491, -0.138] & 3.47 \\
HS pair / Perplexity & Last & 0.40 & 0.073 & -0.608 [-0.777, -0.439] & 3.55 \\
HS pair / Perplexity & First & 0.80 & 0.105 & 2.234 [1.930, 2.538] & 5.91 \\
HS pair / Perplexity & Middle & 0.20 & 0.062 & 2.233 [1.993, 2.473] & 5.54 \\
OpenAI / Jev & Last & 0.27 & 0.019 & 0.929 [0.700, 1.159] & 3.94 \\
OpenAI / Jev & First & 0.47 & 0.025 & 2.790 [2.507, 3.073] & 5.16 \\
OpenAI / Jev & Middle & 0.33 & 0.019 & 2.648 [2.345, 2.952] & 4.68 \\
OpenAI / Clef & Last & 1.00 & 0.041 & -0.335 [-0.677, 0.008] & 3.64 \\
OpenAI / Clef & First & 1.00 & 0.041 & -0.336 [-0.675, 0.002] & 3.37 \\
OpenAI / Clef & Middle & 1.20 & 0.042 & -0.249 [-0.579, 0.080] & 3.51 \\
OpenAI / Perplexity & Last & 0.20 & 0.113 & -3.924 [-4.216, -3.633] & 7.62 \\
OpenAI / Perplexity & First & 0.33 & 0.137 & 0.816 [0.450, 1.183] & 7.83 \\
OpenAI / Perplexity & Middle & 0.07 & 0.092 & 0.338 [0.035, 0.640] & 5.74 \\
\bottomrule
\end{tabular}
\caption{All no-preference placements, n=1,500. Conditional A is P(A)/(P(A)+P(B)) versus baseline; its intervals are descriptive. Non-X flips restrict to items choosing A/B in the X condition. X-first/middle also move the original candidates' absolute positions.}
\label{tab:appendix_tie_placement}
\end{table*}

\begin{table*}[t]
\centering\footnotesize
\setlength{\tabcolsep}{3.5pt}
\begin{tabular}{@{}lrrrr@{}}
\toprule
\shortstack[l]{Dataset\\{}model} & \shortstack[l]{Native reference} & \shortstack[l]{N} & \shortstack[l]{X-last use\\{}\%} & \shortstack[l]{Mean P(X)} \\
\midrule
UKP / Jev & NO\_MAJORITY & 250 & 2.40 & 0.122 \\
UKP / Jev & X & 187 & 22.99 & 0.267 \\
UKP / Clef & NO\_MAJORITY & 250 & 1.60 & 0.105 \\
UKP / Clef & X & 187 & 10.16 & 0.171 \\
UKP / Perplexity & NO\_MAJORITY & 250 & 5.60 & 0.182 \\
UKP / Perplexity & X & 187 & 47.06 & 0.431 \\
HS pair / Jev & ABOUT\_THE\_SAME & 53 & 0.00 & 0.017 \\
HS pair / Clef & ABOUT\_THE\_SAME & 53 & 0.00 & 0.031 \\
HS pair / Perplexity & ABOUT\_THE\_SAME & 53 & 0.00 & 0.093 \\
OpenAI / Jev & SPLIT & 49 & 2.04 & 0.025 \\
OpenAI / Clef & SPLIT & 49 & 2.04 & 0.061 \\
OpenAI / Perplexity & SPLIT & 49 & 2.04 & 0.147 \\
\bottomrule
\end{tabular}
\caption{Non-strict native-reference cases. UKP X is a majority of neither votes; NO\_MAJORITY is separate. HelpSteer ABOUT\_THE\_SAME is published strength zero, and OpenAI SPLIT is equal vote totals. These categories retain their source-specific meanings.}
\label{tab:appendix_tie_reference}
\end{table*}

\begin{table*}[t]
\centering\footnotesize
\setlength{\tabcolsep}{3.5pt}
\begin{tabular}{@{}lrrrr@{}}
\toprule
\shortstack[l]{Model} & \shortstack[l]{Variant} & \shortstack[l]{Score delta\\{}pp [95\% CI]} & \shortstack[l]{Rank $\rho$} & \shortstack[l]{Human MAE\\{}delta pp} \\
\midrule
Jev & Reverse & 2.47 [2.36, 2.57]* & 0.977 & 1.95 \\
Jev & 3 levels & -2.81 [-2.94, -2.69]* & 0.963 & -2.06 \\
Jev & 7 levels & -7.63 [-7.78, -7.48]* & 0.973 & -5.43 \\
Clef & Reverse & -3.86 [-4.07, -3.66]* & 0.984 & -2.80 \\
Clef & 3 levels & -3.55 [-3.77, -3.32]* & 0.984 & -1.79 \\
Clef & 7 levels & -5.19 [-5.41, -4.96]* & 0.978 & -3.57 \\
Perplexity & Reverse & -2.71 [-2.82, -2.60]* & 0.987 & -0.66 \\
Perplexity & 3 levels & -6.78 [-6.89, -6.67]* & 0.988 & -1.04 \\
Perplexity & 7 levels & -3.75 [-3.96, -3.54]* & 0.984 & -1.29 \\
\bottomrule
\end{tabular}
\caption{Humicroedit: remapped scores versus native rubric, n=1,500. * marks 369-test BY. Rank association and saved MAE deltas are descriptive. Reversal changes index polarity; 3/7-level variants also change descriptors. Humor variants additionally change the upper endpoint.}
\label{tab:appendix_formats_humor}
\end{table*}

\begin{table*}[t]
\centering\footnotesize
\setlength{\tabcolsep}{3.5pt}
\begin{tabular}{@{}lrrrr@{}}
\toprule
\shortstack[l]{Model} & \shortstack[l]{Variant} & \shortstack[l]{Score delta\\{}pp [95\% CI]} & \shortstack[l]{Rank $\rho$} & \shortstack[l]{Human MAE\\{}delta pp} \\
\midrule
Jev & Reverse & 0.41 [0.30, 0.51]* & 0.995 & -0.11 \\
Jev & 3 levels & 2.19 [1.97, 2.40]* & 0.990 & 0.83 \\
Jev & 7 levels & 1.36 [1.26, 1.47]* & 0.996 & -0.21 \\
Clef & Reverse & 1.16 [1.02, 1.30]* & 0.980 & 0.25 \\
Clef & 3 levels & 4.69 [4.41, 4.98]* & 0.980 & 3.04 \\
Clef & 7 levels & -1.67 [-1.80, -1.54]* & 0.984 & -0.47 \\
Perplexity & Reverse & -0.38 [-0.48, -0.29]* & 0.995 & 0.12 \\
Perplexity & 3 levels & 0.77 [0.60, 0.94]* & 0.995 & 0.98 \\
Perplexity & 7 levels & 0.76 [0.66, 0.86]* & 0.996 & -0.11 \\
\bottomrule
\end{tabular}
\caption{HelpSteer scores: remapped scores versus native rubric, n=1,500. * marks 369-test BY. Rank association and saved MAE deltas are descriptive. Reversal changes index polarity; 3/7-level variants also change descriptors.}
\label{tab:appendix_formats_helpfulness}
\end{table*}

\begin{table*}[t]
\centering\footnotesize
\setlength{\tabcolsep}{3.5pt}
\begin{tabular}{@{}lrrrr@{}}
\toprule
\shortstack[l]{Model} & \shortstack[l]{Variant} & \shortstack[l]{Score delta\\{}pp [95\% CI]} & \shortstack[l]{Rank $\rho$} & \shortstack[l]{Human MAE\\{}delta pp} \\
\midrule
Jev & Reverse & 1.99 [1.88, 2.11]* & 0.991 & 0.66 \\
Jev & 3 levels & 5.87 [5.64, 6.09]* & 0.982 & 3.97 \\
Jev & 7 levels & 2.58 [2.45, 2.71]* & 0.992 & 0.92 \\
Clef & Reverse & -0.49 [-0.71, -0.26]* & 0.903 & -0.39 \\
Clef & 3 levels & 7.94 [7.63, 8.24]* & 0.908 & 4.78 \\
Clef & 7 levels & -0.72 [-0.87, -0.57]* & 0.938 & -0.13 \\
Perplexity & Reverse & 0.54 [0.45, 0.63]* & 0.990 & -0.22 \\
Perplexity & 3 levels & 1.12 [0.88, 1.36]* & 0.985 & 0.11 \\
Perplexity & 7 levels & 0.67 [0.57, 0.76]* & 0.990 & -0.35 \\
\bottomrule
\end{tabular}
\caption{SummEval relevance: remapped scores versus native rubric, n=1,500. * marks 369-test BY. Rank association and saved MAE deltas are descriptive. Reversal changes index polarity; 3/7-level variants also change descriptors.}
\label{tab:appendix_formats_relevance}
\end{table*}

\begin{table*}[t]
\centering\footnotesize
\setlength{\tabcolsep}{3.5pt}
\begin{tabular}{@{}lrrrrrr@{}}
\toprule
\shortstack[l]{Dataset\\{}model} & \shortstack[l]{Slope} & \shortstack[l]{SD ratio} & \shortstack[l]{Human mid\\{}\%} & \shortstack[l]{Judge mid\\{}\%} & \shortstack[l]{Human\\{}P90-P10} & \shortstack[l]{Judge\\{}P90-P10} \\
\midrule
Humicroedit / Jev & 0.202 & 0.576 & 20.40 & 58.73 & 0.533 & 0.254 \\
Humicroedit / Clef & 0.480 & 0.894 & 20.40 & 39.73 & 0.533 & 0.452 \\
Humicroedit / Perplexity & 0.257 & 0.678 & 20.40 & 38.67 & 0.533 & 0.337 \\
HS score / Jev & 0.462 & 0.981 & 31.80 & 28.73 & 0.600 & 0.555 \\
HS score / Clef & 0.198 & 0.704 & 31.80 & 18.67 & 0.600 & 0.381 \\
HS score / Perplexity & 0.559 & 0.939 & 31.80 & 20.87 & 0.600 & 0.521 \\
SummEval / Jev & 0.418 & 1.096 & 18.27 & 8.40 & 0.344 & 0.347 \\
SummEval / Clef & 0.263 & 0.695 & 18.27 & 6.20 & 0.344 & 0.201 \\
SummEval / Perplexity & 0.220 & 0.620 & 18.27 & 17.27 & 0.344 & 0.192 \\
\bottomrule
\end{tabular}
\caption{Descriptive score compression. Slope regresses judge on human mean; SD ratio is judge/human. Midpoint band is [0.4,0.6]. Dispersion and slopes compare judge scores with the human reference; n=1,500.}
\label{tab:appendix_compression}
\end{table*}

\begin{table*}[t]
\centering\footnotesize
\setlength{\tabcolsep}{3.5pt}
\begin{tabular}{@{}lrr@{}}
\toprule
\shortstack[l]{Dataset\\{}model} & \shortstack[l]{Unadjusted $\rho$} & \shortstack[l]{Adjusted slope\\{}pp per length SD [95\% CI]} \\
\midrule
UKP / Jev & 0.4630 & 10.913 [9.410, 12.417] \\
UKP / Clef & 0.3220 & 8.063 [6.099, 10.027] \\
UKP / Perplexity & 0.4797 & 10.740 [9.359, 12.120] \\
HS pair / Jev & 0.0639 & 0.443 [-1.008, 1.895] \\
HS pair / Clef & 0.3811 & 8.706 [7.399, 10.013] \\
HS pair / Perplexity & 0.1217 & 1.764 [0.544, 2.984] \\
OpenAI / Jev & 0.3481 & 4.005 [2.585, 5.424] \\
OpenAI / Clef & 0.2670 & 3.040 [1.480, 4.601] \\
OpenAI / Perplexity & 0.2007 & -0.543 [-1.756, 0.670] \\
Humicroedit / Jev & 0.0310 & 0.134 [-0.384, 0.652] \\
Humicroedit / Clef & -0.0188 & -0.990 [-1.673, -0.308] \\
Humicroedit / Perplexity & 0.0218 & 0.094 [-0.496, 0.684] \\
HS score / Jev & -0.0173 & -1.216 [-2.303, -0.128] \\
HS score / Clef & 0.1867 & 2.931 [2.098, 3.764] \\
HS score / Perplexity & 0.1404 & 1.380 [0.365, 2.395] \\
SummEval / Jev & 0.1894 & -0.131 [-1.296, 1.033] \\
SummEval / Clef & 0.2553 & 0.645 [-0.183, 1.474] \\
SummEval / Perplexity & 0.3129 & 1.547 [0.944, 2.150] \\
\bottomrule
\end{tabular}
\caption{Observational text-length associations, n=1,500. Pair outcome is P(original A), predictor log(1+words A)-log(1+words B); scoring uses normalized quality and log(1+candidate words). Slopes are per within-dataset predictor SD. Adjustment includes human reference, disagreement, agreement strata and applicable context/category/system covariates. CIs are pointwise descriptive intervals.}
\label{tab:appendix_length}
\end{table*}

\begin{table*}[t]
\centering\footnotesize
\setlength{\tabcolsep}{3.5pt}
\begin{tabular}{@{}lrrrrr@{}}
\toprule
\shortstack[l]{Dataset\\{}model} & \shortstack[l]{Disp. vs\\{}margin $\rho$} & \shortstack[l]{Disp. vs\\{}sensitivity $\rho$} & \shortstack[l]{Disp. vs\\{}repeat SD $\rho$} & \shortstack[l]{Clear abs.\\{}sensitivity pp} & \shortstack[l]{Disputed abs.\\{}sensitivity pp} \\
\midrule
UKP / Jev & -0.1008 & 0.0635 & 0.0956 & 7.31 & 7.76 \\
UKP / Clef & 0.0040 & 0.0278 & 0.0486 & 8.29 & 8.76 \\
UKP / Perplexity & -0.0783 & 0.0971 & 0.0011 & 6.05 & 6.93 \\
HS pair / Jev & -0.2327 & 0.1601 & 0.1841 & 6.51 & 7.81 \\
HS pair / Clef & -0.1052 & 0.0602 & 0.0163 & 6.38 & 6.78 \\
HS pair / Perplexity & -0.2693 & 0.2002 & -0.0117 & 5.04 & 6.32 \\
OpenAI / Jev & -0.5033 & 0.4796 & 0.4688 & 4.23 & 9.44 \\
OpenAI / Clef & -0.3634 & 0.3275 & 0.2307 & 6.43 & 11.40 \\
OpenAI / Perplexity & -0.4737 & 0.3944 & -0.0560 & 4.11 & 6.81 \\
\bottomrule
\end{tabular}
\caption{Descriptive Spearman associations, n=1,500 with 750 per stratum. Dispersion is vote entropy; margin is absolute A/B probability difference. Sensitivity averages absolute changes across three nonbaseline A/B cells.}
\label{tab:appendix_disagreement_pair}
\end{table*}

\begin{table*}[t]
\centering\footnotesize
\setlength{\tabcolsep}{3.5pt}
\begin{tabular}{@{}lrrrrr@{}}
\toprule
\shortstack[l]{Dataset\\{}model} & \shortstack[l]{Disp. vs\\{}prob. var. $\rho$} & \shortstack[l]{Disp. vs\\{}MAE $\rho$} & \shortstack[l]{Disp. vs\\{}repeat SD $\rho$} & \shortstack[l]{Clear abs.\\{}sensitivity pp} & \shortstack[l]{Disputed abs.\\{}sensitivity pp} \\
\midrule
Humicroedit / Jev & -0.0343 & -0.4648 & 0.0044 & 2.30 & 2.23 \\
Humicroedit / Clef & 0.0669 & -0.2645 & -0.0439 & 3.52 & 3.73 \\
Humicroedit / Perplexity & 0.1034 & -0.3196 & 0.0161 & 3.36 & 3.47 \\
HS score / Jev & 0.2807 & -0.0105 & 0.1591 & 1.10 & 1.34 \\
HS score / Clef & 0.1512 & -0.0055 & 0.0358 & 2.32 & 2.55 \\
HS score / Perplexity & 0.3446 & 0.1090 & -0.0623 & 1.13 & 1.39 \\
SummEval / Jev & -0.0472 & 0.0596 & -0.0093 & 1.56 & 1.54 \\
SummEval / Clef & -0.0316 & -0.0723 & -0.0692 & 1.86 & 1.97 \\
SummEval / Perplexity & -0.0276 & -0.1700 & 0.0273 & 0.94 & 1.05 \\
\bottomrule
\end{tabular}
\caption{Descriptive Spearman associations, n=1,500 with 750 per stratum. Dispersion is rating SD/native range (SummEval: mean within-group SD). Probability variance describes the model's rating distribution. Sensitivity averages absolute A-identifier, B-identifier and remapped reverse-anchor changes.}
\label{tab:appendix_disagreement_score}
\end{table*}

\begin{table*}[t]
\centering\footnotesize
\setlength{\tabcolsep}{3.5pt}
\begin{tabular}{@{}lrrrrrr@{}}
\toprule
\shortstack[l]{Model} & \shortstack[l]{Reference} & \shortstack[l]{Mean} & \shortstack[l]{Signed error} & \shortstack[l]{MAE} & \shortstack[l]{Pearson} & \shortstack[l]{Spearman} \\
\midrule
Jev & 3:5 pooled & 0.721 & 0.062 & 0.132 & 0.382 & 0.384 \\
Jev & Experts only & 0.693 & 0.090 & 0.143 & 0.597 & 0.561 \\
Jev & Crowd only & 0.737 & 0.046 & 0.172 & 0.061 & 0.088 \\
Jev & Equal group means & 0.715 & 0.068 & 0.126 & 0.478 & 0.462 \\
Clef & 3:5 pooled & 0.721 & 0.044 & 0.106 & 0.378 & 0.368 \\
Clef & Experts only & 0.693 & 0.072 & 0.135 & 0.592 & 0.556 \\
Clef & Crowd only & 0.737 & 0.028 & 0.143 & 0.060 & 0.071 \\
Clef & Equal group means & 0.715 & 0.050 & 0.102 & 0.473 & 0.450 \\
Perplexity & 3:5 pooled & 0.721 & -0.061 & 0.115 & 0.355 & 0.343 \\
Perplexity & Experts only & 0.693 & -0.033 & 0.140 & 0.559 & 0.526 \\
Perplexity & Crowd only & 0.737 & -0.078 & 0.157 & 0.053 & 0.060 \\
Perplexity & Equal group means & 0.715 & -0.055 & 0.109 & 0.445 & 0.419 \\
\bottomrule
\end{tabular}
\caption{SummEval reference sensitivity on the same 1,500 summaries. The fixed primary reference is the mean of all eight ratings (3 experts, 5 crowd). Alternative references show alignment sensitivity to annotator group and aggregation.}
\label{tab:appendix_reference_population}
\end{table*}

\begin{table*}[t]
\centering\footnotesize
\setlength{\tabcolsep}{3.5pt}
\begin{tabular}{@{}lrrrr@{}}
\toprule
\shortstack[l]{Dataset\\{}model} & \shortstack[l]{Magnitude r\\{}[95\% CI]} & \shortstack[l]{Reader/prior\\{}event N} & \shortstack[l]{Intersect/union\\{}N} & \shortstack[l]{Jaccard} \\
\midrule
UKP / Jev & 0.374 [0.332, 0.414]* & 330 / 74 & 67 / 337 & 0.199 \\
UKP / Clef & 0.380 [0.312, 0.444]* & 152 / 75 & 35 / 192 & 0.182 \\
UKP / Perplexity & 0.396 [0.354, 0.436]* & 606 / 74 & 71 / 609 & 0.117 \\
HS pair / Jev & 0.585 [0.544, 0.624]* & 274 / 64 & 56 / 282 & 0.199 \\
HS pair / Clef & 0.387 [0.323, 0.447]* & 146 / 106 & 56 / 196 & 0.286 \\
HS pair / Perplexity & 0.596 [0.564, 0.625]* & 407 / 98 & 97 / 408 & 0.238 \\
OpenAI / Jev & 0.508 [0.459, 0.553]* & 243 / 50 & 46 / 247 & 0.186 \\
OpenAI / Clef & 0.329 [0.280, 0.376]* & 187 / 100 & 57 / 230 & 0.248 \\
OpenAI / Perplexity & 0.515 [0.476, 0.553]* & 513 / 86 & 85 / 514 & 0.165 \\
Humicroedit / Jev & 0.291 [0.235, 0.346]* & 491 / 93 & 54 / 530 & 0.102 \\
Humicroedit / Clef & 0.348 [0.291, 0.402]* & 911 / 346 & 265 / 992 & 0.267 \\
Humicroedit / Perplexity & 0.753 [0.729, 0.775]* & 1430 / 1472 & 1425 / 1477 & 0.965 \\
HS score / Jev & 0.889 [0.877, 0.900]* & 1453 / 1351 & 1344 / 1460 & 0.921 \\
HS score / Clef & 0.207 [0.143, 0.269]* & 476 / 220 & 108 / 588 & 0.184 \\
HS score / Perplexity & 0.735 [0.704, 0.763]* & 1407 / 1320 & 1291 / 1436 & 0.899 \\
SummEval / Jev & 0.622 [0.574, 0.666]* & 899 / 747 & 610 / 1036 & 0.589 \\
SummEval / Clef & 0.155 [0.099, 0.210]* & 343 / 102 & 33 / 412 & 0.080 \\
SummEval / Perplexity & 0.531 [0.484, 0.574]* & 1421 / 1499 & 1420 / 1500 & 0.947 \\
\bottomrule
\end{tabular}
\caption{Reader/prior-judge susceptibility across all 18 model/dataset combinations, selected by cue identity. Pearson associations use absolute within-item contrast magnitudes and disjoint source calls. Events are choice reversals (pairs) or absolute normalized score shifts >0.05 (scores). * marks 1,656-test BY; dependence-aware Fisher-z pointwise CIs.}
\label{tab:appendix_shared_vulnerability}
\end{table*}

\begin{table*}[t]
\centering\footnotesize
\setlength{\tabcolsep}{3.5pt}
\begin{tabular}{@{}lrrrr@{}}
\toprule
\shortstack[l]{Model\\{}contrast} & \shortstack[l]{Pooled pp\\{}[95\% CI]} & \shortstack[l]{Q} & \shortstack[l]{I$^2$\\{}\%} & \shortstack[l]{$\tau$$^2$\\{}pp$^2$} \\
\midrule
Jev / First-position & -3.53 [-9.03, 1.97] & 60.86 & 97.2 & 4.740 \\
Jev / B-label & 0.57 [-6.60, 7.75] & 1610.19 & 99.9 & 8.342 \\
Jev / Numeric & 5.35 [-1.68, 12.37] & 117.66 & 98.0 & 7.838 \\
Jev / Human-AI & 8.02 [-4.93, 20.96] & 931.96 & 99.9 & 27.091 \\
Jev / Reader & 13.35 [12.04, 14.65]* & 5.91 & 67.1 & 0.180 \\
Jev / Prior & 1.97 [-1.70, 5.65] & 337.62 & 99.4 & 2.170 \\
Jev / Inject first & 0.45 [-1.53, 2.44] & 137.58 & 98.7 & 0.628 \\
Jev / Inject second & -0.31 [-1.47, 0.86] & 45.36 & 95.5 & 0.210 \\
Clef / First-position & -0.07 [-0.29, 0.16] & 1.32 & 0.0 & 0.000 \\
Clef / B-label & -1.71 [-23.60, 20.19] & 612.88 & 99.8 & 77.418 \\
Clef / Numeric & -1.58 [-21.60, 18.43] & 498.54 & 99.7 & 64.648 \\
Clef / Human-AI & 11.82 [-32.15, 55.79] & 3094.14 & 100.0 & 313.035 \\
Clef / Reader & 5.44 [1.78, 9.11] & 88.32 & 98.2 & 2.124 \\
Clef / Prior & 2.07 [0.86, 3.28] & 21.88 & 89.5 & 0.212 \\
Clef / Inject first & 1.69 [-1.79, 5.16] & 214.32 & 98.8 & 1.929 \\
Clef / Inject second & 0.99 [-1.36, 3.34] & 67.43 & 97.5 & 0.867 \\
Perplexity / First-position & -0.05 [-11.17, 11.07] & 655.38 & 99.7 & 19.973 \\
Perplexity / B-label & -1.09 [-2.54, 0.36] & 91.44 & 97.4 & 0.331 \\
Perplexity / Numeric & -1.21 [-12.82, 10.41] & 646.93 & 99.7 & 21.798 \\
Perplexity / Human-AI & 6.60 [0.02, 13.17] & 696.50 & 99.7 & 6.980 \\
Perplexity / Reader & 23.15 [12.25, 34.04] & 279.15 & 99.5 & 19.072 \\
Perplexity / Prior & 4.04 [1.95, 6.13] & 148.83 & 98.6 & 0.699 \\
Perplexity / Inject first & 2.77 [0.60, 4.94] & 131.71 & 98.5 & 0.750 \\
Perplexity / Inject second & 6.23 [-3.42, 15.89] & 1033.30 & 99.8 & 15.069 \\
\bottomrule
\end{tabular}
\caption{All choice cross-task pools: three tasks each, REML between-task variance and modified Knapp-Hartung t inference (df=2); Q df=2. I2 uses REML-consistent typical sampling variance. * marks 1,656-test BY. CIs are pointwise; pooling summarizes the three selected tasks. Weights and all leave-one-task-out results are retained electronically.}
\label{tab:appendix_meta_choice}
\end{table*}

\begin{table*}[t]
\centering\footnotesize
\setlength{\tabcolsep}{3.5pt}
\begin{tabular}{@{}lrrrr@{}}
\toprule
\shortstack[l]{Model\\{}contrast} & \shortstack[l]{Pooled pp\\{}[95\% CI]} & \shortstack[l]{Q} & \shortstack[l]{I$^2$\\{}\%} & \shortstack[l]{$\tau$$^2$\\{}pp$^2$} \\
\midrule
Jev / B-A & 0.11 [-1.54, 1.75] & 572.89 & 99.7 & 0.438 \\
Jev / Numeric & 0.15 [-1.81, 2.10] & 760.95 & 99.8 & 0.619 \\
Jev / Human-AI & 0.76 [-1.28, 2.79] & 540.03 & 99.7 & 0.667 \\
Jev / Reader & 8.69 [-8.17, 25.55] & 4322.35 & 100.0 & 46.026 \\
Jev / Prior & 6.47 [-5.40, 18.34] & 4809.01 & 100.0 & 22.811 \\
Jev / Inject & 1.35 [-3.37, 6.07] & 896.84 & 99.8 & 3.599 \\
Jev / Reverse & 1.62 [-1.06, 4.31] & 851.46 & 99.7 & 1.166 \\
Clef / B-A & 0.11 [-0.78, 1.00] & 42.48 & 95.6 & 0.123 \\
Clef / Numeric & 0.06 [-0.26, 0.38] & 5.95 & 65.1 & 0.011 \\
Clef / Human-AI & 0.73 [-1.13, 2.60] & 181.70 & 98.9 & 0.556 \\
Clef / Reader & 4.57 [-1.36, 10.51] & 412.46 & 99.6 & 5.676 \\
Clef / Prior & 1.50 [-0.10, 3.11] & 104.96 & 98.1 & 0.410 \\
Clef / Inject & -1.36 [-6.38, 3.66] & 513.88 & 99.5 & 4.062 \\
Clef / Reverse & -1.06 [-7.42, 5.29] & 1587.85 & 99.9 & 6.535 \\
Perplexity / B-A & 0.12 [-2.70, 2.95] & 1574.38 & 99.8 & 1.293 \\
Perplexity / Numeric & -0.12 [-0.41, 0.16] & 23.29 & 94.6 & 0.012 \\
Perplexity / Human-AI & 0.25 [-3.16, 3.67] & 926.88 & 99.9 & 1.881 \\
Perplexity / Reader & 12.79 [-5.72, 31.29] & 4576.10 & 100.0 & 55.441 \\
Perplexity / Prior & 12.88 [-0.61, 26.36] & 3186.90 & 100.0 & 29.442 \\
Perplexity / Inject & 0.82 [-3.93, 5.58] & 1537.18 & 99.9 & 3.665 \\
Perplexity / Reverse & -0.85 [-5.01, 3.30] & 2178.99 & 99.9 & 2.798 \\
\bottomrule
\end{tabular}
\caption{All score cross-task pools: three tasks each, REML between-task variance and modified Knapp-Hartung t inference (df=2); Q df=2. I2 uses REML-consistent typical sampling variance. No pooled contrast meets the 1,656-test BY threshold. CIs are pointwise; pooling summarizes the three selected tasks. Weights and all leave-one-task-out results are retained electronically.}
\label{tab:appendix_meta_score}
\end{table*}

\begin{table*}[t]
\centering\footnotesize
\setlength{\tabcolsep}{3.5pt}
\begin{tabular}{@{}lrrrr@{}}
\toprule
\shortstack[l]{Model\\{}UKP contrast} & \shortstack[l]{Dyadic\\{}pp [95\% CI]} & \shortstack[l]{Component\\{}pp [95\% CI]} & \shortstack[l]{Bootstrap\\{}pp [95\% CI]} & \shortstack[l]{Omit largest\\{}pp} \\
\midrule
Jev / First-position & -2.250 [-2.977, -1.524] & -2.250 [-3.147, -1.354] & [-3.223, -1.456] & -2.60 \\
Jev / B-label & 3.856 [3.655, 4.058] & 3.856 [3.646, 4.066] & [3.668, 4.090] & 3.88 \\
Jev / Human vs AI & 7.029 [6.696, 7.361] & 7.029 [6.727, 7.330] & [6.725, 7.321] & 7.03 \\
Jev / Reader high vs low & 13.545 [12.956, 14.135] & 13.545 [12.897, 14.194] & [12.972, 14.223] & 13.73 \\
Clef / First-position & 0.015 [-0.258, 0.288] & 0.015 [-0.253, 0.283] & [-0.250, 0.293] & 0.06 \\
Clef / B-label & -8.643 [-9.620, -7.665] & -8.643 [-9.564, -7.721] & [-9.647, -7.752] & -8.65 \\
Clef / Human vs AI & 3.568 [3.072, 4.063] & 3.568 [3.151, 3.984] & [3.140, 3.992] & 3.55 \\
Clef / Reader high vs low & 4.775 [4.364, 5.185] & 4.775 [4.398, 5.151] & [4.390, 5.170] & 4.76 \\
Perplexity / First-position & 4.582 [4.064, 5.100] & 4.582 [4.097, 5.066] & [4.105, 5.094] & 4.62 \\
Perplexity / B-label & -1.498 [-1.717, -1.279] & -1.498 [-1.714, -1.282] & [-1.723, -1.287] & -1.50 \\
Perplexity / Human vs AI & 9.021 [8.620, 9.421] & 9.021 [8.547, 9.494] & [8.593, 9.533] & 9.16 \\
Perplexity / Reader high vs low & 27.799 [26.783, 28.815] & 27.799 [26.637, 28.960] & [26.712, 28.994] & 28.10 \\
\bottomrule
\end{tabular}
\caption{Dependence sensitivity for all models and four UKP headline contrasts; 2,000 bootstrap draws with a fixed seed. The largest connected joke group has 207 pairs. Results are sensitivity diagnostics.}
\label{tab:appendix_dependence}
\end{table*}

\begin{table*}[t]
\centering\footnotesize
\setlength{\tabcolsep}{3.5pt}
\begin{tabular}{@{}lrr@{}}
\toprule
\shortstack[l]{Check} & \shortstack[l]{Quantity} & \shortstack[l]{Alternative estimate} \\
\midrule
Baseline snapshot & Jev: UKP position (pp) & -2.24 \\
Five-call mean & Jev: UKP position (pp) & -2.23 \\
Score reconstruction & Jev: Helpfulness reader (pp) & 16.52 \\
Band reweighting & Jev: Humor MAE (pp) & 25.93 \\
Lower-length subset & Jev: UKP position (pp) & -2.23 [-3.01, -1.46] \\
Component variance & Jev: UKP position (pp) & -2.25 [-3.15, -1.35] \\
\bottomrule
\end{tabular}
\caption{Representative numerical sensitivity checks, all from existing diagnostics. Jev's primary UKP position effect is -2.25 [-2.98, -1.52] pp; primary helpfulness reader effect is 16.47 [16.12, 16.81] pp; primary humor MAE is 25.56 pp. Baseline alternatives address the factorial; reweighting addresses baseline score error. Length and component intervals are sensitivity diagnostics. Full cases and exceptions are summarized in the accompanying text and retained in the electronic tables.}
\label{tab:appendix_robustness_summary}
\end{table*}

\end{document}